\documentclass{article}

\usepackage{microtype}
\usepackage{graphicx}
\usepackage{subcaption}
\usepackage{booktabs} 
\usepackage{multirow}   
\usepackage{array}      
\usepackage{enumitem}
\usepackage{hyperref}

\newcounter{jsonformat}
\renewcommand{\thejsonformat}{\arabic{jsonformat}}
\newcommand{\jsoncaption}[1]{%
  \vspace{1pt}\par\noindent
  \footnotesize\textbf{Format~\thejsonformat.}~#1\par
}

\usepackage[preprint]{icml2026}

\usepackage{amsmath}
\usepackage{amssymb}
\usepackage{mathtools}
\usepackage{amsthm}
\usepackage{pifont}
\newcommand{\cmark}{\ding{51}} 

\usepackage[capitalize,noabbrev]{cleveref}

\theoremstyle{plain}

\theoremstyle{definition}

\theoremstyle{remark}

\usepackage{tcolorbox}
\tcbuselibrary{skins,breakable}
\usepackage[textsize=tiny]{todonotes}

\usepackage{xcolor}
\usepackage[table]{xcolor}  
\usepackage{tcolorbox}
\tcbuselibrary{skins}
\definecolor{jsonbg}{HTML}{F7F7F9}      
\definecolor{jsonkey}{HTML}{2C5282}     
\definecolor{jsonval}{HTML}{2F855A}     
\definecolor{jsonframe}{HTML}{E2E8F0}   

\definecolor{TopOneColor}{RGB}{0,60,150}     
\definecolor{TopTwoColor}{RGB}{60,120,200}   

\newcommand{\TopOne}[1]{\textbf{\textcolor{TopOneColor}{#1}}}
\newcommand{\TopTwo}[1]{\textbf{\textcolor{TopTwoColor}{#1}}}

\newtcolorbox{jsonbox}[1][]{
    enhanced,
    colback=jsonbg,
    colframe=jsonframe,
    boxrule=0.8pt,          
    arc=3mm,                
    left=8pt, right=8pt, top=6pt, bottom=6pt,
    fontupper=\ttfamily\small, 
    halign=left,            
}

\newcommand{\jkey}[1]{\textcolor{jsonkey}{\textbf{"#1"}}} 
\newcommand{\jval}[1]{\textcolor{jsonval}{#1}}

\icmltitlerunning{StableI2I: Spotting Unintended Changes in Image-to-Image Transition}

\begin{document}

\twocolumn[
  \icmltitle{StableI2I: Spotting Unintended Changes in Image-to-Image Transition}



\icmlsetsymbol{equal}{*}
\icmlsetsymbol{equal_c}{\textdagger}
\icmlsetsymbol{intern}{+}

\begin{icmlauthorlist}
    \icmlauthor{Jiayang Li}{equal,pku,shailab,intern}
    \icmlauthor{Shuo Cao}{equal,shailab,ustc}
    \icmlauthor{Xiaohui Li}{shailab,sjtu}
    \icmlauthor{Zhizhen Zhang}{pku}
    \icmlauthor{Kaiwen Zhu}{shailab,sjtu}
    \icmlauthor{Yule Duan}{pku}
    \icmlauthor{Yu Qiao}{shailab}
    \icmlauthor{Jian Zhang}{equal_c,pku}
    \icmlauthor{Yihao Liu}{equal_c,shailab}
\end{icmlauthorlist}

\icmlaffiliation{shailab}{Shanghai Artificial Intelligence Laboratory}
\icmlaffiliation{pku}{Peking University}
\icmlaffiliation{ustc}{University of Science and Technology of China}
\icmlaffiliation{sjtu}{Shanghai Jiao Tong University}

\icmlcorrespondingauthor{Jian Zhang}{zhangjian.sz@pku.edu.cn}
\icmlcorrespondingauthor{Yihao Liu}{liuyihao@pjlab.org.cn}

  \icmlkeywords{Machine Learning, ICML}

  \vskip 0.3in
]



 \printAffiliationsAndNotice{
\icmlEqualContribution
\textsuperscript{+} Work done during an internship at Shanghai Artificial Intelligence Laboratory.
}

\begin{abstract}
In most real-world image-to-image (I2I) scenarios, existing evaluations primarily focus on instruction following and the perceptual quality or aesthetics of the generated images. However, they largely fail to assess whether the output image preserves the semantic correspondence and spatial structure of the input image. To address this limitation, we propose StableI2I, a unified and dynamic evaluation framework that explicitly measures content fidelity and pre--post consistency across a wide range of I2I tasks without requiring reference images, including image editing and image restoration. In addition, we construct StableI2I-Bench, a benchmark designed to systematically evaluate the accuracy of MLLMs on such fidelity and consistency assessment tasks. Extensive experimental results demonstrate that StableI2I provides accurate, fine-grained, and interpretable evaluations of content fidelity and consistency, with strong correlations to human subjective judgments. Our framework serves as a practical and reliable evaluation tool for diagnosing content consistency and benchmarking model performance in real-world I2I systems. The project page and source code are publicly available at \url{https://henry-lee-real.github.io/StableI2I_Page}.

\end{abstract}

\section{Introduction}

\begin{figure}[t] 
  \centering 
  \vspace{-5pt}
  \includegraphics[width=\columnwidth]{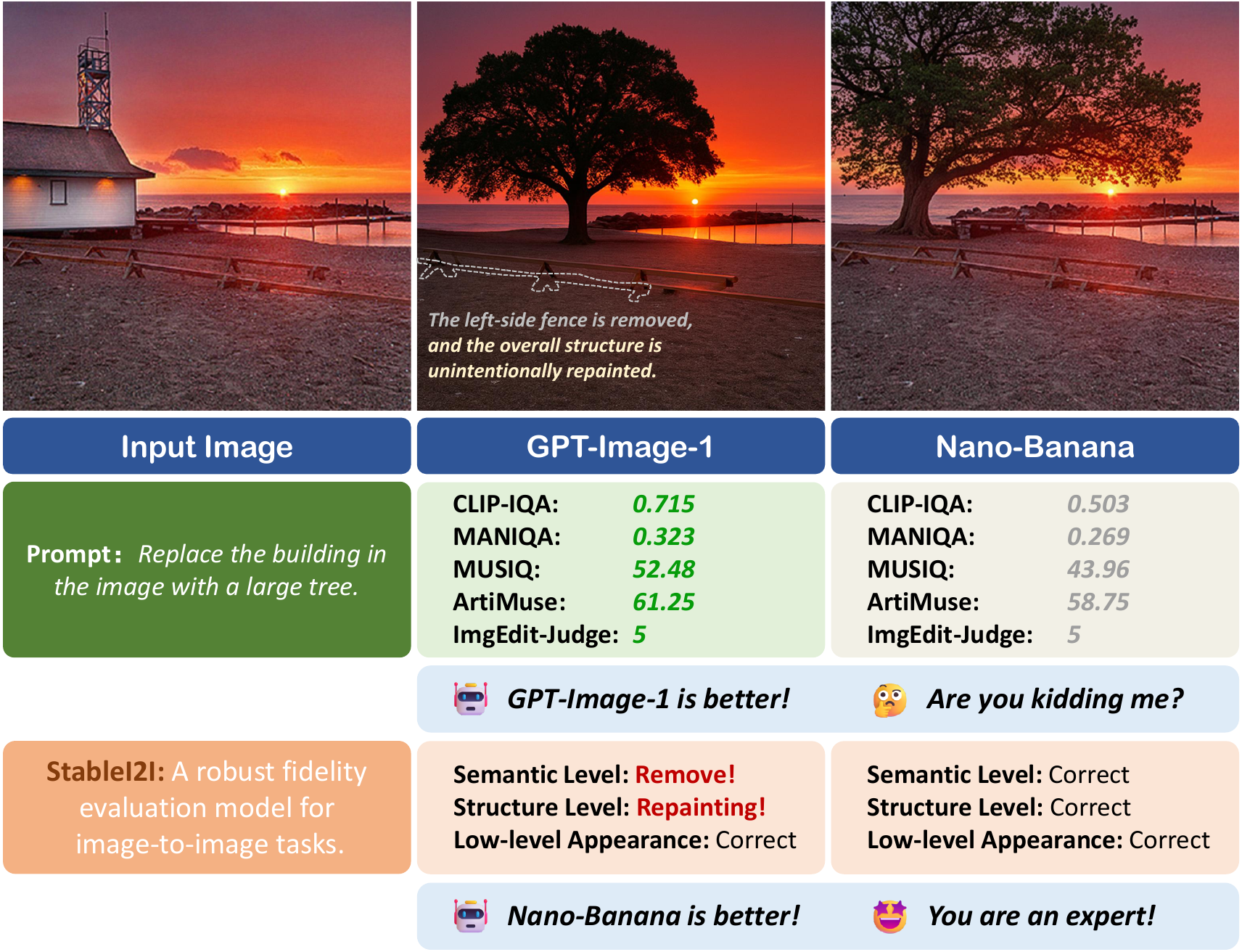} 
  \captionsetup{skip=1pt}
  \caption{Qualitative image editing results from GPT-Image-1 and Nano-Banana, with scores from multiple evaluation metrics. CLIP-IQA~\cite{clipiqa}, MANIQA~\cite{maniqa}, and MUSIQ~\cite{musiq} are conventional IQA metrics, while ArtiMuse~\cite{artimuse} is a recent IAA metric. ImgEdit-Judge~\cite{imgedit} reports scores under the Physical \& Detail Coherence dimension. In contrast, StableI2I more accurately assesses content fidelity and consistency.}
  \vspace{-25pt}
  \label{fig:intro}
\end{figure}


With the rapid advancement of generative models~\cite{flux,controlnet}, current systems are increasingly capable of following user instructions and producing high-quality images. However, the inherent randomness of the sampling process often leads to substantial information drift between the generated output and the input image. Even state-of-the-art models such as Nano-Banana~\cite{google_gemini_image} are affected by this issue. This phenomenon highlights the urgent need for effective methods to evaluate and calibrate content drift.


However, current I2I evaluations mainly focus on instruction following and output aesthetics~\cite{artimuse} or perceptual quality~\cite{clipiqa}, while largely ignoring whether the output image remains faithful to the input image during editing or restoration (Fig.~\ref{fig:intro}).
Although images generated by GPT-Image-1 achieve higher scores under existing metrics, their texture and semantic content still exhibit unintended changes, including unnecessary repainting of the sky and sandy areas, and the disappearance of the left-side fence relative to the input image.
Without explicitly assessing pre--post consistency, such inconsistencies can lead to severe consequences in high-stakes I2I applications, such as medical imaging and remote sensing.
Therefore, a principled evaluation method is required to jointly consider the input image, output image, and processing instruction to assess content fidelity before and after transformation.

For image editing tasks, a commonly adopted strategy~\cite{ryu2025towards} is to use a mask to separate the edited region and then compare the remaining areas for consistency.
However, valid edits often give rise to necessary global variations, such as changes in illumination, shadows, or other secondary effects that are causally induced by the edit itself.
For example, in the output images of Fig.~\ref{fig:intro}, after the object is replaced with a tree, the shadow cast beneath it is a reasonable and physically plausible outcome.
In such cases, rigid mask-based separation becomes inappropriate and can easily lead to erroneous judgments.
Moreover, mask-based methods are not applicable to image restoration tasks, where the entire image may be altered.
Consequently, an effective I2I evaluation framework must be capable of understanding the editing instruction, interpreting image content, and dynamically producing analysis results conditioned on both.

Recent studies~\cite{Step1X-Edit} have also recognized this limitation and attempted to address it by leveraging prompt engineering to query powerful proprietary MLLMs for consistency judgments. Although current closed-source MLLMs exhibit strong semantic-level image understanding capabilities, they remain insensitive to fine-grained pixel-level and structural information~\cite{unipercept}. As a result, such evaluation methods often produce cases where semantic content appears consistent while pixel-level content is misaligned. As shown in Fig.~\ref{fig:intro}, ImgEdit-Judge assigns an incorrect score under the \emph{Physical \& Detail Coherence} dimension, failing to detect substantial content repainting. This deficiency arises because ImgEdit-Judge is distilled from the closed-source GPT-4o model~\cite{gpt4} and lacks explicit sensitivity to pixel-level structure.



Motivated by these observations, we propose \textbf{StableI2I}, a fidelity-oriented I2I evaluation model that jointly considers semantic and pixel-level consistency. By integrating these dimensions, StableI2I better judges semantic content and pixel-level details between the input and output images.


StableI2I adapts to different I2I tasks by conditioning on the input instruction and selectively attending to regions and attributes that must remain consistent.
We further define three complementary fidelity dimensions: Structure Level, Semantic Level, and Low-level Appearance.
In addition, we introduce \textbf{StableI2I-Bench}, a benchmark with formatted question–answer pairs for systematically evaluating modern MLLMs on I2I fidelity assessment across these three dimensions, reflecting both high-level semantic reasoning and low-level visual perception.
We also propose an \emph{error-amplification} data construction pipeline to mitigate the long-tail distribution of subtle consistency violations.
In summary, our main contributions are as follows:

\begin{itemize}
\vspace{-10pt}
\item We propose \textbf{StableI2I}, a fidelity-oriented evaluation model for I2I tasks that jointly captures semantic-level and pixel-level consistency.
\vspace{-5pt}
\item We introduce \textbf{StableI2I-Bench}, a benchmark designed to assess models' integrated high-level and low-level visual reasoning abilities for fidelity evaluation.
\vspace{-5pt}
\item We develop a multi-stage, multi-task data construction pipeline that enhances data diversity and improves the robustness of model capabilities.
\vspace{-5pt}
\end{itemize}

\section{Related Works}

\subsection{Quality Assessment for I2I Transition}
Quality assessment models for natural image transition are conventionally classified into Full-Reference (FR) and No-Reference (NR) paradigms~\cite{psnr,lpips,fid,clipiqa,clipscore,qalign,deqa_score,artimuse,unipercept}.
While FR metrics~\cite{pieapp,dists} rely on ground truths that are often unavailable, standard NR methods predominantly evaluate absolute aesthetics or perceptual quality~\cite{qalign,unipercept}, failing to capture the semantic consistency with the source input that is essential for image editing.
This limitation motivates a source-conditioned evaluation paradigm that explicitly accounts for content fidelity and structural preservation in the absence of ground truth.

\subsection{MLLM-based I2I Transition Assessment}
Evaluating I2I transition requires a multi-dimensional perspective that encompasses semantic consistency and aesthetic quality, yet this critical domain remains largely under-explored.
Prior works~\cite{imgedit,Step1X-Edit,instructscore,PartEdit} primarily rely on general-purpose MLLMs, either through prompt engineering as in MagicBrush~\cite{magicbrush} and CompBench~\cite{compbench}, or via distillation methods such as ImgEdit~\cite{imgedit}, which trains a judge using GPT-4o~\cite{gpt4} priors without specific adaptation for I2I transition.
Consequently, these approaches are predominantly coarse-grained and biased toward high-level semantic consistency, often failing to capture low-level pixel-wise variations or provide professional-grade diagnostic depth.
These limitations highlight the need for a fidelity-centric and instruction-aware evaluation framework that jointly considers both semantic and perceptual consistency.

\begin{figure*}[t] 
  \centering 
  \includegraphics[width=\textwidth]{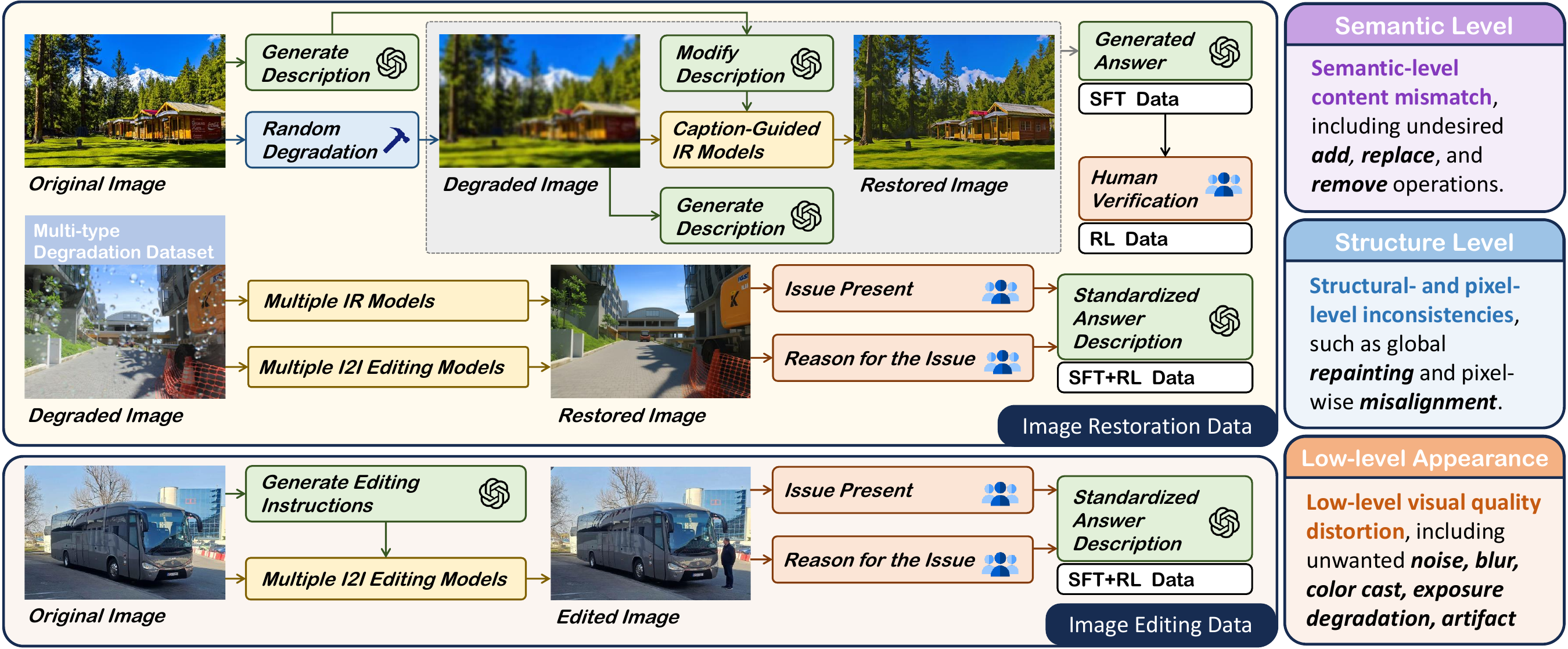} 
  \captionsetup{skip=1.5pt}
  \caption{The data construction pipeline mainly includes data construction for image editing tasks and image restoration tasks, with annotations provided along three dimensions shown on the right.}
  \vspace{-18pt}
  \label{fig:data_pipeline}
\end{figure*}

\section{From Data to Benchmark and Training}
\subsection{Data Construction Pipeline}
\label{subsec:31}
\vspace{-5pt}
I2I tasks can be broadly categorized into two types: high-level semantic editing and low-level image restoration.
Because these two task types emphasize different objectives, most existing models tend to focus primarily on either high-level semantics or low-level perceptual quality, while paying insufficient attention to the other, which often leads to fidelity issues.
\textbf{For image editing}, models focus on preserving and modifying object-level content, which makes it difficult to maintain low-level texture details. As a result, many existing models exhibit unintended content repainting and pixel-level mismatches in regions that should remain unchanged, even though object-level semantics are preserved. \textbf{For image restoration}, models may not truly understand what object should be restored, i.e., they lack sufficient semantic capability, which leads to semantic drift in the restored content.

To address these issues, we design an error-amplification data generation pipeline together with a corresponding annotation pipeline.
As shown in Fig.~\ref{fig:data_pipeline}, for the image restoration task, we first apply random degradations to collected natural images. We then use GPT-5 to extract faithful content descriptions of the original images and introduce controlled semantic perturbations to deliberately alter and corrupt these descriptions.
The corrupted descriptions are subsequently used to guide a text-guided image restoration model for restoration.
In this way, the restoration model is guided by incorrect semantic information and is forced to restore the low-quality image toward an incorrect semantic direction, which significantly increases the probability of generating erroneous samples.
For the image editing task, since it is difficult to deliberately construct erroneous data through a deterministic pipeline, we generate diverse editing results using multiple types of editing instructions together with multiple generative models.
The specific models, data sources, and dataset scales used in our data pipeline are detailed in Appendix~\ref{sec:A1}.

Based on the above I2I data, we define three categories of error types, as illustrated in Fig.~\ref{fig:data_pipeline}:
\textbf{Semantic Level}: whether unintended additions, deletions, or modifications occur in semantic content that should be preserved;
\textbf{Structure Level}: whether the output image exhibits texture or structural misalignment relative to the input image, or unintended content repainting;
\textbf{Low-level Appearance}: whether the output image exhibits low-level degradations relative to the input image, such as noise, blur, color shift, or artifacts.
With these three fidelity dimensions defined, we annotate two types of data, as illustrated in Fig.~\ref{fig:data_pipeline}.
For the image restoration task, we adopt a semi-automatic annotation scheme: for pipeline-synthesized data, since the corrupted semantic information is known by construction, we use the GPT-5 API for first-stage automatic annotations, followed by human filtering and correction; for restoration results obtained under real-world settings, we rely on fully manual annotation to label all fidelity-related content.
For all data from the image editing task, we also employ fully manual annotation to label the complete content.
Details of the number of annotators and the annotator training procedure are provided in Appendix~\ref{sec:A2}.

\subsection{StableI2I-Bench: Benchmark Definition}
\vspace{-5pt}
Most existing I2I tasks rely on prompt engineering to let closed-source models evaluate the pre–post consistency of I2I results. To assess whether existing open-source and closed-source models can truly use prompts to correctly judge consistency, we release StableI2I-Bench.

We randomly sample 1,000 human-annotated image pairs from each of the three dimensions—Semantic Level, Structure Level, and Low-level Appearance—to construct the benchmark. The benchmark adopts a formatted prompt design, where each prompt includes the input image, the output image, the I2I control instruction, background knowledge describing the evaluation dimension, and a specification of the required output format, together with the corresponding structured answers. The detailed prompt templates and benchmark examples are provided in Appendix~\ref{sec:A3}.

\begin{figure*}[!ht] 
  \centering 
  \includegraphics[width=\textwidth]{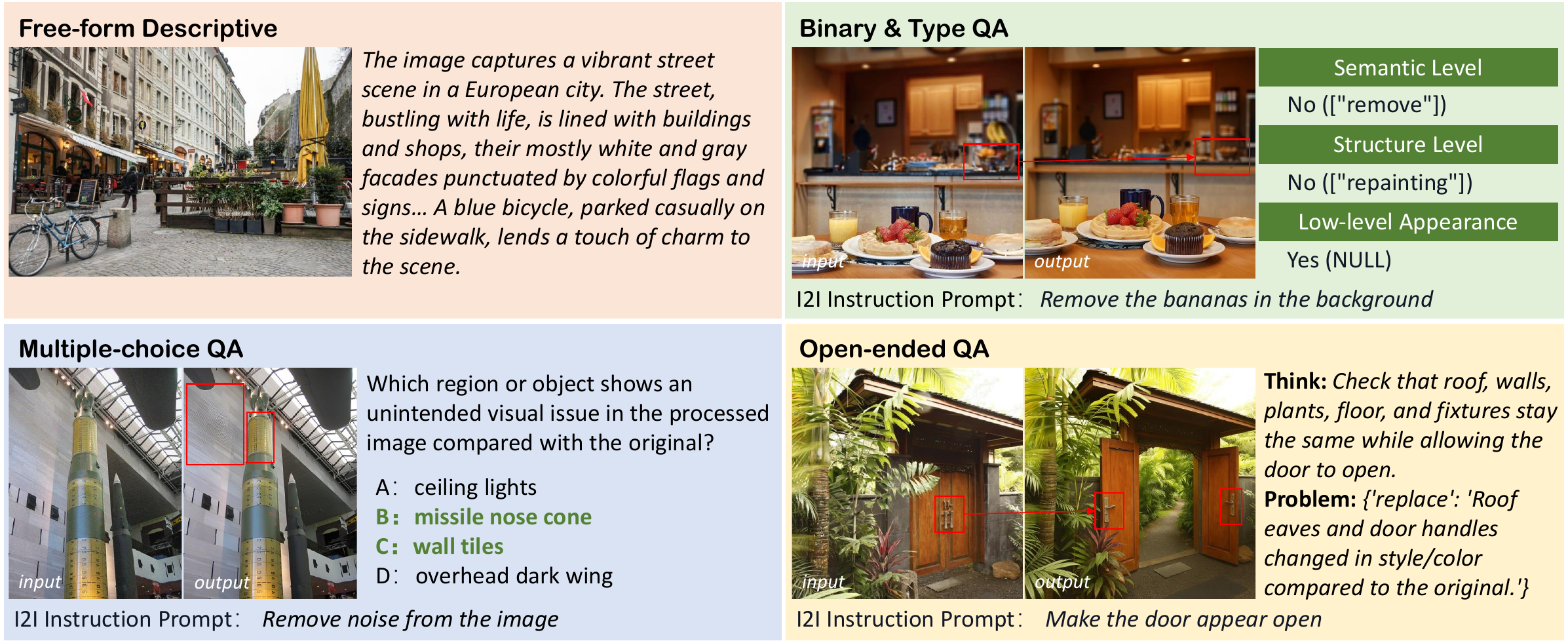} 
  \captionsetup{skip=1.5pt}
  \caption{An illustration of the four types of training data: Free-form Descriptive, Binary \& Type QA, Multiple-choice QA, and Open-ended QA.}
  \vspace{-18pt}
  \label{fig:training_data_type}
\end{figure*}

\subsection{StableI2I-Train: Training Corpus Construction}
\vspace{-5pt}
Since StableI2I is fine-tuned on a relatively small 8B-parameter MLLM~\cite{qwen3vl}, and given the limited model capacity at this scale, we adopt fixed task templates during training to ensure stable and reliable evaluation behavior.
We first define two fundamental data types:
\textbf{Binary \& Type QA}, which produces concise and standardized evaluation outputs following Format~\ref{fig:output-format}, and
\textbf{Open-ended QA}, which provides detailed natural-language descriptions of observed errors, with output structures corresponding to Format~\ref{fig:detailed-output-format}.
Concrete examples of both output formats are shown in Fig.~\ref{fig:training_data_type}, and the fixed input templates for these two task types are provided in Appendix~\ref{sec:A42}.

{%
\setlength{\parskip}{0pt}%
\par\vspace*{0pt}
\centering
\resizebox{0.9\linewidth}{!}{%
\begin{jsonbox}
\refstepcounter{jsonformat}
\label{fig:output-format}
    \{ \jkey{answer}: \jval{Yes} \text{ or } \jval{No}, \ 
       \jkey{problem}: \jval{Null} \text{ or } [\jval{Type$_1$}, \jval{Type$_2$}, \dots] \}
\end{jsonbox}
}
\vspace*{-2pt}
\jsoncaption{Unified output format for Binary \& Type QA.}
\par
}%

{%
\setlength{\parskip}{0pt}%
\par\vspace*{2pt}
\centering
\resizebox{0.9\linewidth}{!}{%
\begin{jsonbox}
\refstepcounter{jsonformat}
\label{fig:detailed-output-format}
    \{ 
       \jkey{think}: \jval{CoT}, \ 
       \jkey{problem}: \{ 
           \jkey{Type$_1$}: \jval{Detail$_1$}, 
           \jkey{Type$_2$}: \jval{Detail$_2$}, 
           \dots 
       \} 
    \}
\end{jsonbox}}
\vspace*{-2pt}
\jsoncaption{Unified output format for Open-ended QA.}
\par
}%

In addition, to preserve the model’s basic visual perception and descriptive abilities, we introduce a multi-task descriptive QA dataset termed \textbf{Free-form Descriptive}, as illustrated in Fig.~\ref{fig:training_data_type}.
This data is mainly sourced from ShareGPT4V~\cite{sharegpt4v} and CapRL~\cite{caprl}, covering diverse modalities and content types, including natural images, AIGC images, tables, and multiple QA styles such as descriptive and multiple-choice formats.

\begin{figure}[h] 
  \centering 
  \vspace{-5pt}
  \includegraphics[width=\columnwidth]{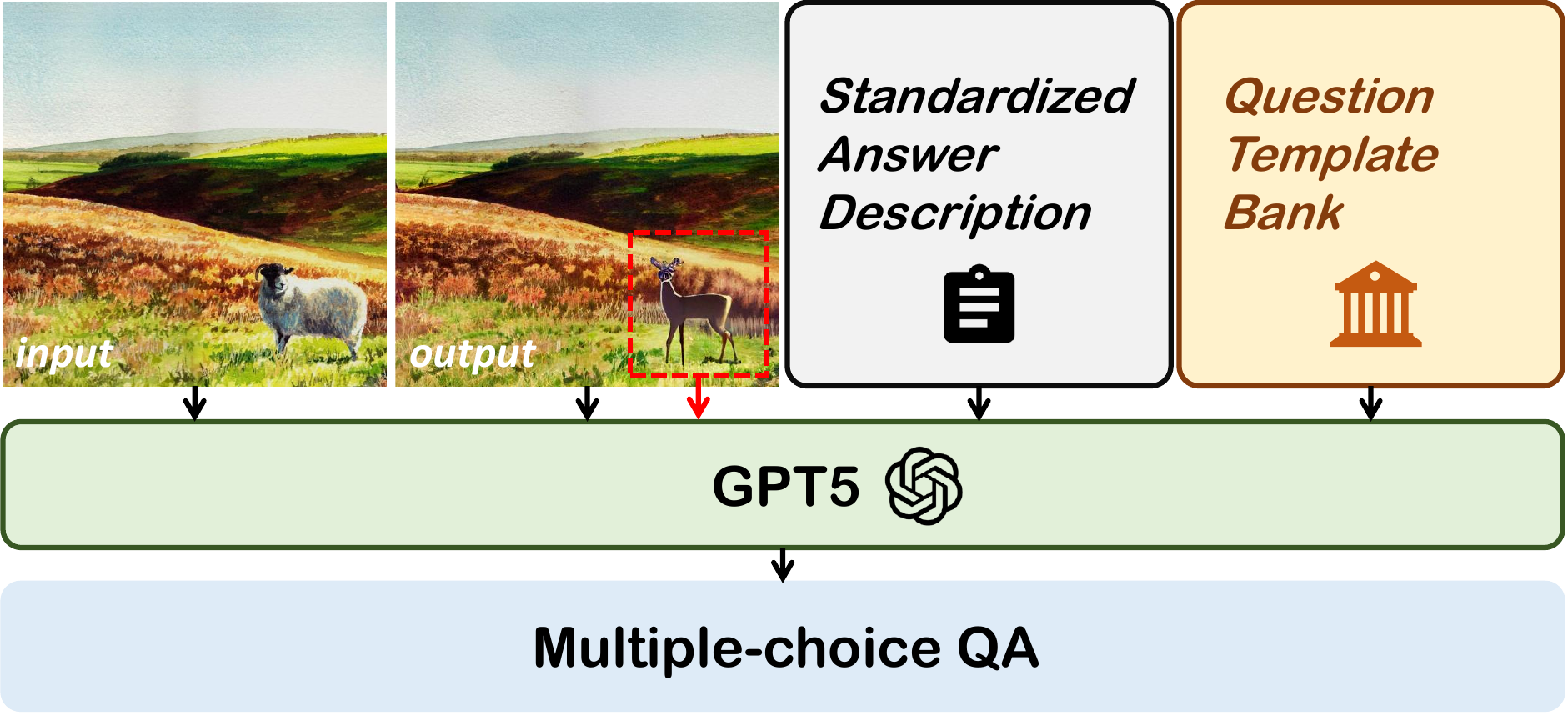} 
  \captionsetup{skip=1pt}
  \caption{Pipeline for Constructing the Multiple-Choice QA Dataset.}
  \vspace{-15pt}
  \label{fig:mcp}
\end{figure}

As our training framework incorporates reinforcement learning to improve generalization, the Open-ended QA data introduces practical challenges.
Its free-form outputs are difficult to constrain using structured reward functions, making it only feasible to reliably evaluate fixed output formats and coarse-grained content correctness.
To address this issue, we reorganize human-annotated descriptions into \textbf{Multiple-choice QA}, as shown in Fig.~\ref{fig:mcp}.
This conversion transforms open-ended descriptions into deterministic choice-based questions, enabling the model to improve its fine-grained content understanding and analytical ability by selecting the correct options.
More details on the construction of Multiple-choice QA are provided in Appendix~\ref{sec:A41}, and representative QA examples are shown in Fig.~\ref{fig:training_data_type}.

However, this strategy alone is far from sufficient.
As discussed in Section~\ref{subsec:31} (Data Construction Pipeline), it is difficult to construct and annotate large-scale I2I editing data that contains diverse and realistic errors.
In the next Section~\ref{sec:model}, we will describe in detail how we expand the data scale and enhance model capability through a multi-stage training scheme.

In addition, the existing data scale remains insufficient to effectively improve the weak pixel-level perceptual capability of the ViT encoder.
We therefore introduce Texture-Aware Enhancement Data to enhance the encoder’s perception at the pixel level.
Details of its construction pipeline and the data composition of the overall StableI2I-Train dataset are provided in Appendix~\ref{sec:A41}.


\begin{figure*}[t] 
  \centering 
  \includegraphics[width=0.8\textwidth]{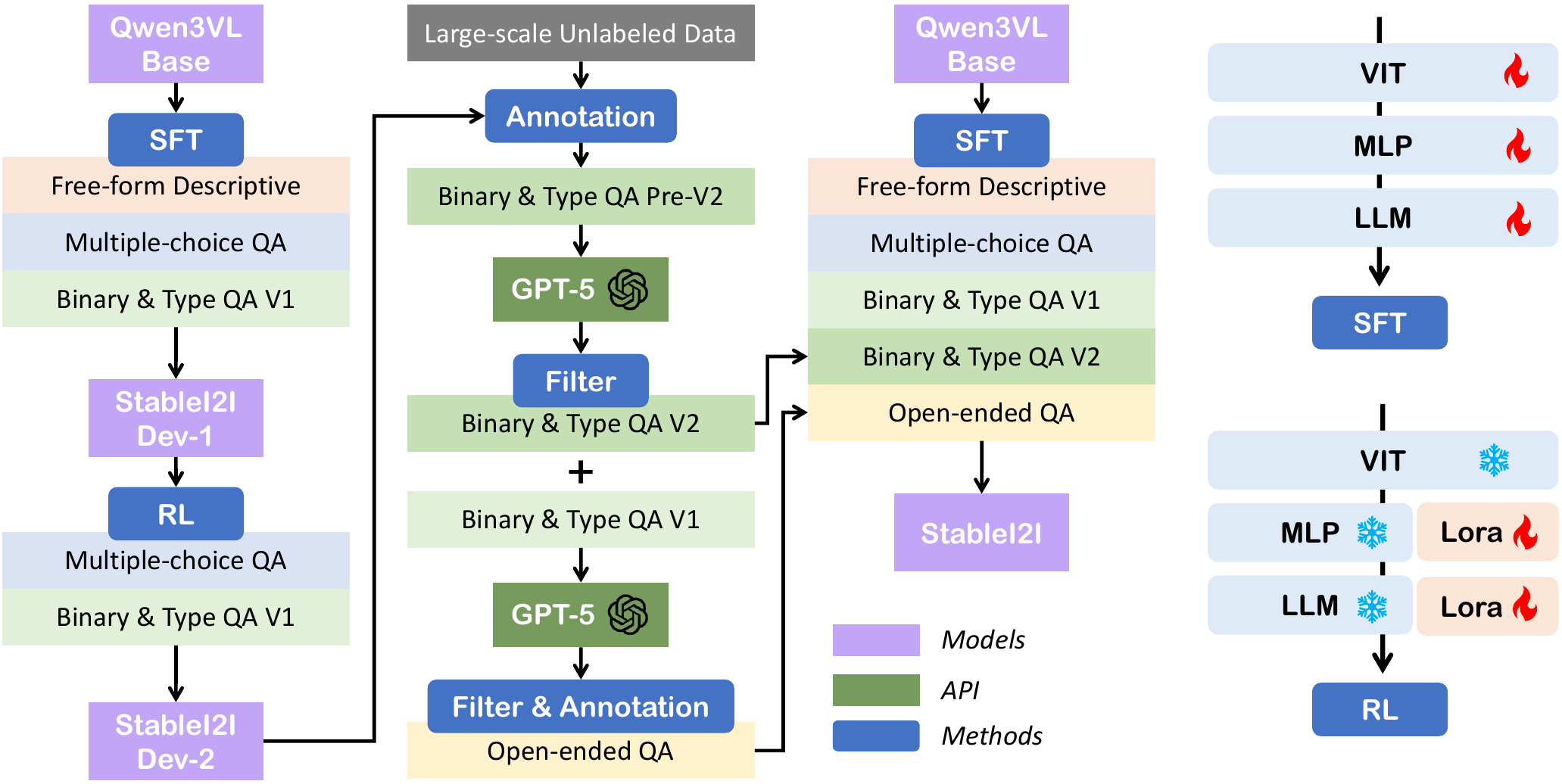} 
  \captionsetup{skip=1.5pt}
  \caption{The three columns on the left illustrate the training pipeline, including the training strategy at different stages and the corresponding data composition. The single column on the right shows the configuration of trainable model parameters under different training strategies.}
  \vspace{-18pt}
  \label{fig:training_pipeline}
\end{figure*}

\section{StableI2I}
\label{sec:model}
\vspace{-5pt}

For training, we first perform supervised fine-tuning on Qwen3-VL-8B-Instruct~\cite{qwen3vl} using Binary \& Type QA, Multiple-choice QA, and Free-form Descriptive data, enabling the model to perform basic task responses while preserving visual perception and comprehension abilities. We then apply reinforcement learning with GRPO on the SFT-trained model to further improve generalization. The rewards are defined separately for Multiple-Choice (MC) tasks corresponding to Multiple-choice QA data, and Binary Answer tasks corresponding to Binary \& Type QA data.

For MC tasks, let \(G\) and \(\hat{G}\) denote the ground-truth and predicted option sets. If \(\hat{G} \nsubseteq G\), the reward is zero; otherwise,
$R_{\mathrm{MC}} = M \cdot \frac{|\hat{G}|}{|G|},$
where \(M\) is the maximum MC reward.

For Binary Answer tasks, each output contains an \textit{answer} and a \textit{problem} field (see Format~(\ref{fig:output-format})). We first require the predicted \textit{answer} to exactly match the ground truth; otherwise, the reward is zero. When the ground-truth answer is \emph{Yes}, a reward of 1 is assigned only if both problem sets are empty. When the answer is \emph{No}, both problem sets must be non-empty. Let \(P\) and \(\hat{P}\) denote the ground-truth and predicted problem type sets. The reward is computed as
\vspace{-3pt}
\begin{equation}
R_{\mathrm{Binary}} =
\max\!\left(
0,\;
\frac{|\hat{P} \cap P|}{|P|}
- \alpha \frac{|\hat{P} \setminus P|}{|P|}
\right),
\label{eq:reward-binary}
\end{equation}
\vspace{-5pt}
where \(\alpha\) penalizes false positive predictions.

\begin{figure*}[t]
\centering
\noindent
\begin{minipage}[t]{0.63\textwidth}
\vspace{0pt}

\captionsetup{type=table}
\caption{Quantitative comparison of mainstream models on StableI2I-Bench.
Binary Accuracy measures answer correctness, while Strict Accuracy additionally requires correct error types.
Best and second-best results are highlighted in \TopOne{dark blue} and \TopTwo{light blue}, respectively.}
\vspace{-6pt}
\label{tab:cmp_model}

\scriptsize
\setlength{\tabcolsep}{3.5pt}
\renewcommand{\arraystretch}{1.05} 

\scalebox{0.95}{%

\begin{tabular}{
l
@{\hspace{4pt}\vrule width 0.4pt\hspace{4pt}}
*{4}{>{\centering\arraybackslash}c}
@{\hspace{4pt}\vrule width 0.4pt\hspace{4pt}}
*{4}{>{\centering\arraybackslash}c}
}
\toprule
\multirow{2}{*}{\textbf{Models}}
& \multicolumn{4}{c@{\hspace{4pt}\vrule width 0.4pt\hspace{4pt}}}{\textbf{Binary Accuracy}}
& \multicolumn{4}{c}{\textbf{Strict Accuracy}} \\
\cmidrule(lr){2-5} \cmidrule(lr){6-9}
& Structure & Semantic & Low-level & Avg.
& Structure & Semantic & Low-level & Avg. \\
\midrule

\rowcolor{blue!10}
\multicolumn{9}{l}{\textbf{Open-Source Models}} \\

Qwen3VL-8B-Instruct  
& 36.60 & 55.60 & 81.60 & 57.93
& 13.80 & 31.70 & 63.60 & 36.37 \\

Qwen3VL-32B-Instruct 
& 53.20 & 73.60 & 87.70 & 71.50
& 34.90 & 48.90 & 56.30 & 46.70 \\

InternVL-3.5-8B      
& 36.30 & 64.60 & 59.10 & 53.33
& 13.00 & 21.30 & 28.10 & 20.80 \\

InternVL-3.5-38B     
& 50.10 & 64.90 & 81.70 & 65.57
& 42.30 & 39.40 & 31.60 & 37.77 \\

\midrule

\rowcolor{gray!12}
\multicolumn{9}{l}{\textbf{Proprietary Models}} \\

Grok-4.1               
& 50.50 & 73.70 & 77.90 & 67.37
& 38.50 & 56.30 & 28.70 & 41.17 \\

Claude-Sonnet-4.5      
& 66.20 & 70.10 & 89.70 & 75.33
& 62.40 & 54.40 & 73.30 & 63.37 \\

Claude-Sonnet-4.5-think
& 63.80 & 69.40 & 84.70 & 72.63
& \TopTwo{62.50} & 56.20 & 65.20 & 61.30 \\

Gemini-2.5-pro         
& 66.67 & 79.90 & 90.70 & 79.09
& 56.66 & 58.60 & 37.20 & 50.82 \\

Gemini-3-pro           
& \TopTwo{71.52} & \TopOne{83.61} & 91.72 & \TopTwo{82.28}
& 62.19 & \TopTwo{63.69} & \TopTwo{75.56} & \TopTwo{67.15} \\

GPT-4o                 
& 59.90 & 79.70 & \TopTwo{94.70} & 78.10
& 46.60 & 60.80 & 71.10 & 59.50 \\

GPT-5                  
& 65.50 & \TopTwo{83.00} & 93.20 & 80.57
& 54.60 & 60.20 & 51.00 & 55.27 \\

\midrule

StableI2I              
& \TopOne{85.40} & 82.80 & \TopOne{99.10} & \TopOne{89.10}
& \TopOne{83.70} & \TopOne{67.30} & \TopOne{98.00} & \TopOne{83.00} \\

\bottomrule
\end{tabular}

} 
\end{minipage}
\hfill
\noindent
\begin{minipage}[t]{0.35\textwidth}
\vspace{0pt}
\centering

\includegraphics[width=\linewidth]{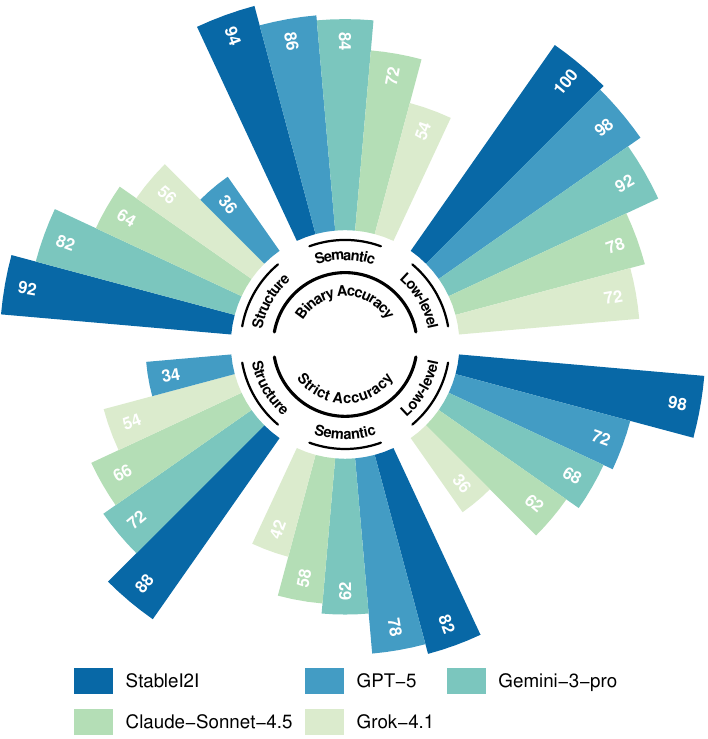}

\captionsetup{type=figure}
\vspace{-4pt}
\caption{Human evaluation of answer accuracy.}
\label{fig:human}
\end{minipage}

\vspace{-15pt}
\end{figure*}

We then use the reinforcement learning–trained model to annotate large-scale unlabeled data from various I2I tasks. Subsequently, GPT-5 is employed to filter out samples with obvious errors. Next, we combine the original Binary \& Type QA data with the newly annotated Binary \& Type QA data to construct Open-ended QA data, corresponding to Format~(\ref{fig:detailed-output-format}), where GPT-5 generates the associated \textit{think} and \textit{problem} fields. Finally, we perform full fine-tuning of Qwen3-VL-8B-Instruct using the mixture of all original and newly annotated data to obtain the final model. Through this training paradigm, the model achieves stronger perceptual understanding and improved generalization capability. The overall training pipeline is illustrated in Fig.~\ref{fig:training_pipeline}.

\section{Experiments}
\vspace{-5pt}
\textbf{Training Details.} We adopt two types of training paradigms in our experiments: supervised fine-tuning (SFT) and reinforcement learning (RL). For SFT, we use a learning rate of \(1 \times 10^{-5}\) with a cosine learning rate scheduler and a warmup ratio of 0.03. The total batch size is set to 128, and the model is trained for 5 epochs over the full training set. For RL, we also use a learning rate of \(1 \times 10^{-5}\) together with a cosine scheduler and a warmup ratio of 0.03. A weight decay of 0.01 is applied for regularization. We employ GRPO as the policy optimization algorithm, where 16 candidate responses are generated for each training sample. During training, 8 samples are processed in parallel. The RL experiments are run for at least 5{,}000 training steps. Unless otherwise specified, both SFT and RL experiments are conducted using 8 NVIDIA H200 GPUs.

\subsection{Evaluation Results on StableI2I-Bench}
\vspace{-5pt}
Here, we evaluate a range of mainstream open-source and proprietary multimodal large models on our StableI2I-Bench to analyze whether these models can accurately assess fidelity in I2I tasks.
For open-source models, we select Qwen3VL-8B-Instruct~\cite{qwen3vl}, Qwen3VL-32B-Instruct~\cite{qwen3vl}, InternVL-3.5-8B~\cite{internvl3}, and InternVL-3.5-38B~\cite{internvl3}.
For proprietary models, we include Grok-4.1~\cite{xai_grok4_1}, Claude-Sonnet-4.5~\cite{anthropic_claude_sonnet}, Claude-Sonnet-4.5-think~\cite{anthropic_claude_sonnet}, Gemini-2.5-pro~\cite{gemini}, Gemini-3-pro~\cite{gemini}, GPT-4o~\cite{gpt4}, and GPT-5~\cite{openai_gpt5}.

The evaluation results are reported in Tab.~\ref{tab:cmp_model}. 
And we note an important detail regarding the evaluation setting. 
The input template used by StableI2I at inference time is not identical to the template provided in StableI2I-Bench for evaluating general-purpose MLLMs. 
However, both settings use exactly the same image pairs $(I_{\text{in}}, I_{\text{out}})$ and the same I2I control instruction $x$. 
This discrepancy arises because StableI2I is a specialized model trained for I2I fidelity assessment, 
and its training relies on a fixed task template. 
As a result, compared to general-purpose MLLMs, StableI2I exhibits weaker robustness to template variations 
when prompts contain additional prior knowledge or more complex instruction structures.

To assess the impact of template priors, we additionally report in Appendix~\ref{sec:B1} 
the results of several mainstream MLLMs evaluated under the simplified StableI2I template. 
The results show that, after removing the structured priors explicitly provided in the benchmark template, 
the performance of general-purpose models drops significantly across all three fidelity dimensions 
and becomes substantially worse than their performance under the original benchmark template.

As shown in Tab.~\ref{tab:cmp_model}, among all mainstream models, Gemini-3-pro achieves the best overall performance. 
In general, these models perform strongest at the Semantic Level, 
which is likely because most contemporary models primarily focus on high-level visual information. 
Moreover, when prominent low-level degradations are present in the output image, 
models are often able to respond to such issues to some extent. 
In contrast, most models perform relatively poorly on Structure Level QA. 
This can be mainly attributed to two factors. 
First, the task requires pixel-level alignment. 
Second, in some samples, although the semantic content remains consistent, 
the global structure has been repainted or altered.
After training and fine-tuning, our model outperforms existing state-of-the-art vision models on this task. 
This result further indicates that there remains substantial room for improvement 
in current visual models with respect to I2I fidelity assessment.

To verify that our evaluation results align with human priors, we recruited seven volunteers to assess the correctness of the model outputs shown in Fig.~\ref{fig:human}.
Specifically, we randomly selected 50 images generated by Bagel, Nano-Banana and GPT-Image-1 on ImgEdit-Bench, and asked the model to perform evaluations using the response templates defined in StableI2I-Bench.
The results indicate that the evaluation outputs produced by StableI2I are largely consistent with human judgments. In addition, we provide supplementary quantitative comparisons with ImgEdit-Judge~\cite{imgedit} in the Appendix.~\ref{sec:app}.

\begin{table*}[t]
\centering
\caption{Quantitative results of mainstream I2I models on image editing and restoration tasks evaluated using StableI2I. The reported values correspond to the accuracy of each evaluation dimension over the entire benchmark. The best-performing results are highlighted in \TopOne{dark blue}, while the second-best results are highlighted in \TopTwo{light blue}.}
\vspace{-5pt}
\label{tab:performance}
\scriptsize
\setlength{\tabcolsep}{5pt} 
\renewcommand{\arraystretch}{1.12}

\resizebox{\textwidth}{!}{%
\begin{tabular}{
l
@{\hspace{3pt}\vrule width 0.4pt\hspace{3pt}}
*{4}{>{\centering\arraybackslash}c}
@{\hspace{3pt}\vrule width 0.4pt\hspace{3pt}}
*{4}{>{\centering\arraybackslash}c}
@{\hspace{3pt}\vrule width 0.4pt\hspace{3pt}}
*{4}{>{\centering\arraybackslash}c}
}
\toprule
\multirow{2}{*}{\textbf{Datasets}}
& \multicolumn{4}{c@{\hspace{3pt}\vrule width 0.4pt\hspace{3pt}}}{\textbf{ImgEdit-Bench}}
& \multicolumn{4}{c@{\hspace{3pt}\vrule width 0.4pt\hspace{3pt}}}{\textbf{GEdit-Bench}}
& \multicolumn{4}{c}{\textbf{Low-level Dataset}} \\
\cmidrule(lr){2-5}\cmidrule(lr){6-9}\cmidrule(lr){10-13}
& Semantic & Structure & Low-level & Avg.
& Semantic & Structure & Low-level & Avg.
& Semantic & Structure & Low-level & Avg. \\
\midrule

\rowcolor{blue!10}
\multicolumn{13}{l}{\textbf{Open-Source Models}} \\

Lumina-DiMOO
& 0.9366 & 0.2465 & 0.8732 & 0.6854
& 0.7913 & 0.0776 & 0.5677 & 0.4790
& 0.6880 & 0.2740 & 0.4910 & 0.4843 \\

Flux.1-dev
& 0.3345 & 0.0123 & 0.9701 & 0.4390
& 0.2368 & 0.0223 & 0.8589 & 0.3727
& 0.2400 & 0.1140 & 0.4590 & 0.2710 \\

OmniGen2
& 0.8803 & 0.6567 & 0.6655 & 0.7342
& 0.8325 & \TopTwo{0.6881} & 0.7294 & 0.7518
& 0.8320 & 0.6600 & 0.5260 & 0.6727 \\

Bagel
& \TopOne{0.9718} & \TopOne{0.8750} & 0.8046 & \TopOne{0.8838}
& 0.8870 & \TopOne{0.8003} & 0.7979 & \TopOne{0.8292}
& \TopOne{0.9520} & \TopOne{0.9240} & \TopOne{0.5630} & \TopOne{0.8130} \\

Qwen-Image-Edit-2509
& 0.9525 & \TopTwo{0.6849} & \TopTwo{0.9718} & \TopTwo{0.8697}
& \TopTwo{0.9068} & 0.6271 & \TopTwo{0.9142} & \TopTwo{0.8174}
& 0.8480 & 0.6620 & 0.5390 & 0.6830 \\

Qwen-Image-Edit-2511
& 0.9595 & 0.4683 & 0.9349 & 0.7876
& \TopOne{0.9134} & 0.5899 & 0.8977 & 0.8021
& 0.8720 & 0.6620 & 0.5450 & 0.6930 \\

\midrule

\rowcolor{gray!12}
\multicolumn{13}{l}{\textbf{Proprietary Models}} \\

GPT-Image-1
& 0.8390 & 0.1342 & \TopOne{0.9839} & 0.6524
& 0.6160 & 0.0693 & \TopOne{0.9182} & 0.5347
& 0.7333 & 0.0283 & 0.4717 & 0.4111 \\

Nano-Banana
& \TopTwo{0.9665} & 0.6772 & 0.9506 & 0.8648
& 0.8803 & 0.5070 & 0.8908 & 0.7594
& \TopTwo{0.8878} & \TopTwo{0.7455} & \TopTwo{0.5591} & \TopTwo{0.7308} \\

\bottomrule

\end{tabular}%
}
\vspace{-10pt}
\end{table*}

\begin{figure*}[!h] 
  \centering 
  \includegraphics[width=\textwidth]{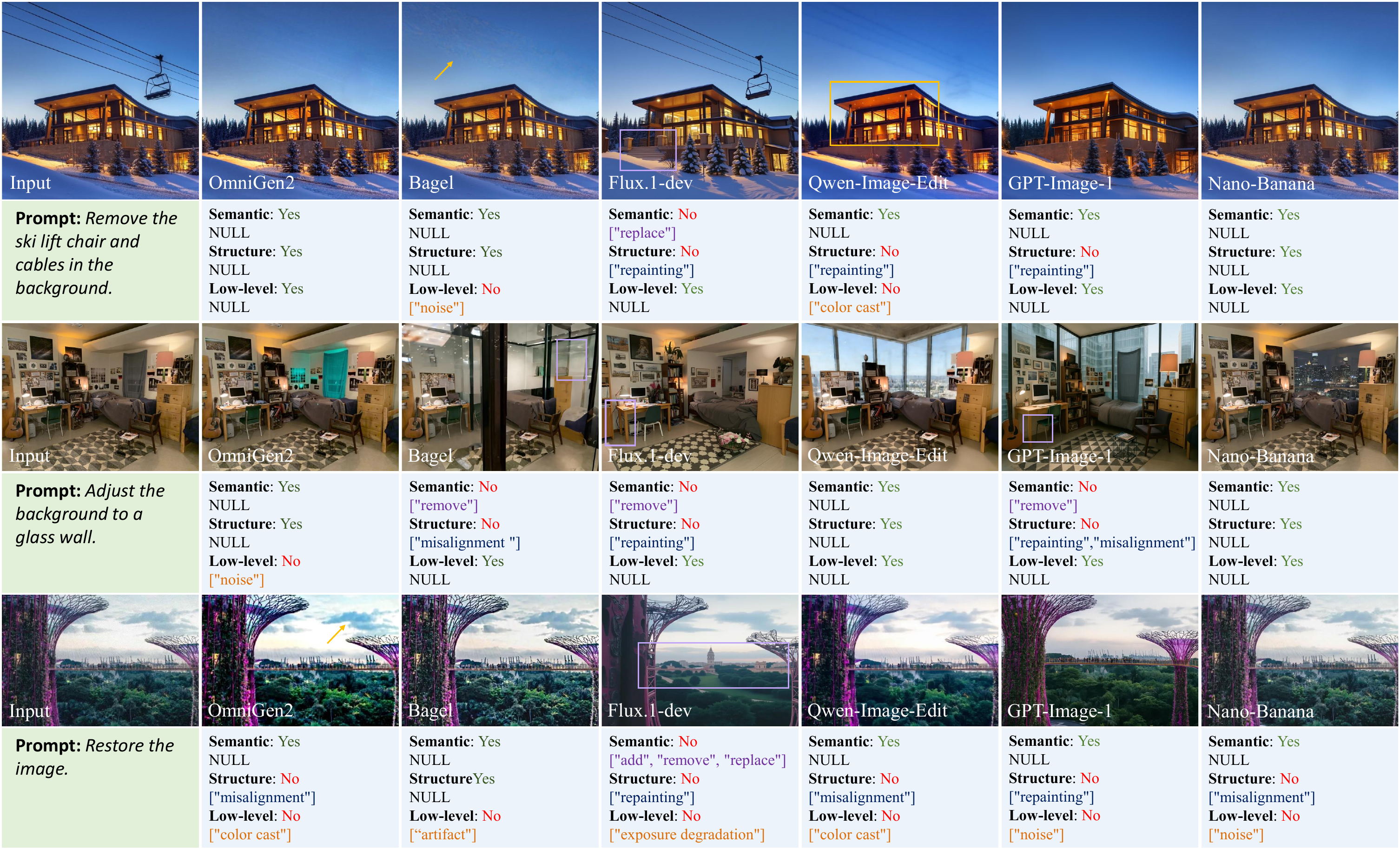} 
  \captionsetup{skip=1.5pt}
  \caption{Qualitative results of mainstream I2I models on image editing and restoration tasks evaluated using StableI2I. From top to bottom, the three groups of examples are drawn from ImgEdit-Bench, GEdit-Bench, and the Low-level Dataset, respectively. Qwen-Image-Edit refers to the Qwen-Image-Edit-2511 model release. For each evaluation dimension in StableI2I, an output of ``Yes'' indicates no detected error, whereas ``No'' denotes the presence of an error, accompanied by a brief problem type describing the corresponding inconsistency. Please zoom in for better visualization of fine-grained details.}
  \vspace{-18pt}
  \label{fig:cmp_i2i}
\end{figure*}

\subsection{Model Performance on I2I Tasks Assessed by StableI2I}
\vspace{-5pt}
In this section, we use StableI2I to score the performance of mainstream generative models on multiple existing image editing benchmarks as well as on a collection of image restoration tasks.
For image editing benchmarks, we adopt ImgEdit-Bench~\cite{imgedit} and GEdit-Bench~\cite{Step1X-Edit}.
For low-level restoration tasks, we construct a dataset by sampling data from various scenarios, including denoising, deblurring, deraining, dehazing, and exposure correction. 
See Appendix~\ref{sec:A1} for more details.

The open-source models evaluated include Lumina-DiMOO~\cite{lumina-dimoo}, Flux.1-dev~\cite{flux}, OmniGen2~\cite{omnigen2}, Bagel~\cite{bagel}, Qwen-Image-Edit-2509~\cite{Qwen-Image}, and Qwen-Image-Edit-2511~\cite{Qwen-Image},
while the proprietary models include GPT-Image-1~\cite{Qwen-Image} and Nano-Banana~\cite{google_gemini_image}.

The scoring protocol measures the proportion of samples for which StableI2I answers ``Yes" (i.e., no fidelity issues detected) in each evaluation dimension, relative to the total number of samples.
Detailed results are reported in Tab.~\ref{tab:performance}.

From the above table, we find that Bagel and the Qwen series demonstrate superior performance in terms of fidelity. In general, most models achieve acceptable results at the semantic level, suggesting that they can largely preserve high-level semantic information. In contrast, their performance at the structure level is markedly worse, indicating a limited ability to maintain fine-grained structural consistency. 
Notably, Flux.1-dev and GPT-Image-1 both suffer from severe content repainting in real-world evaluations, where the original structural layout is largely discarded and regenerated. This behavior leads to extremely low Structure Level scores for both models.

\begin{figure*}[!h] 
  \centering 
  \includegraphics[width=\textwidth]{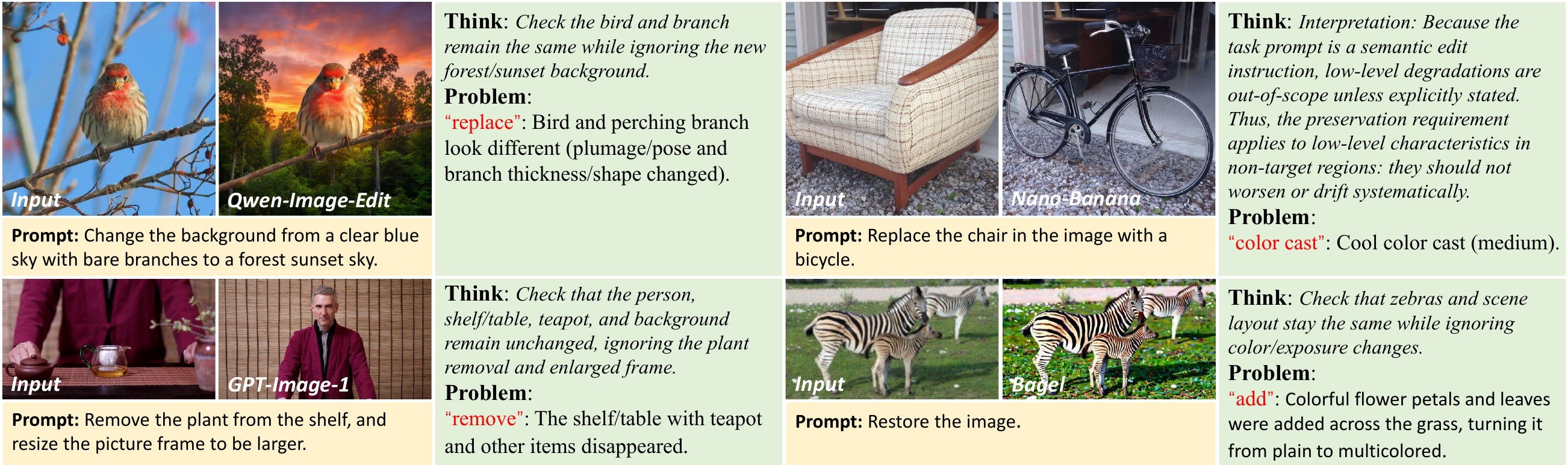} 
  \captionsetup{skip=1.5pt}
  \caption{This figure presents representative failure cases of different models on ImgEdit-Bench, together with a detailed analysis of the observed errors. Zooming in is recommended for better visualization of fine-grained details.}
  \vspace{-18pt}
  \label{fig:des}
\end{figure*}

Fig.~\ref{fig:cmp_i2i} provides a qualitative overview of StableI2I’s evaluations across image editing and restoration tasks. We observe that all three types of information drift can occur in real-world scenarios, with unintended content repainting being the most critical issue for current generative models.
Fig.~\ref{fig:des} further illustrates StableI2I’s detailed error descriptions. Since structural errors correspond to global changes with limited categories, we focus on a finer-grained analysis of Semantic Level and Low-level Appearance, along with their associated error types.

\subsection{Ablation Study}
\vspace{-7pt}
\begin{table}[h]
\centering
\caption{The following shows a quantitative ablation study on the use of Multiple-Choice QA. The reported values are the accuracy of samples where both the answer and the problem type in Binary \& Type QA are predicted correctly.
}
\vspace{-5pt}
\label{tab:ab1}
\tiny
\setlength{\tabcolsep}{4.5pt}
\renewcommand{\arraystretch}{0.85}
\setlength{\extrarowheight}{-0.3pt}
\resizebox{\columnwidth}{!}{%
\begin{tabular}{
@{} 
*{2}{>{\centering\arraybackslash}c}
@{\hspace{2pt}\vrule width 0.4pt\hspace{2pt}}
*{2}{>{\centering\arraybackslash}c}
@{\hspace{2pt}\vrule width 0.4pt\hspace{2pt}}
*{4}{>{\centering\arraybackslash}c}
@{} 
}
\toprule

\multicolumn{2}{c@{\hspace{3pt}\vrule width 0.4pt\hspace{3pt}}}{\textbf{SFT}}
& \multicolumn{2}{c@{\hspace{3pt}\vrule width 0.4pt\hspace{3pt}}}{\textbf{RL}}
& \multicolumn{4}{c}{\textbf{Binary \& Type QA}} \\
\cmidrule(lr){1-2}\cmidrule(lr){3-4}\cmidrule(lr){5-8}

w/o & w
& w/o & w
& Structure & Semantic & Low-level & Avg. \\
\midrule
 & 
&  & 
& 53.50 & 21.40 & 24.60 & 33.17 \\

 & 
& \cmark & 
& 63.90 & 65.10 & 65.80 & 64.95 \\

\cmark & 
& \cmark & 
& 79.50 & 67.80 & 94.10 & 80.47 \\

\cmark & 
&  & \cmark
& 81.10 & 67.10 & 94.00 & 80.73 \\

 & \cmark
& & \cmark
& 81.80 & 68.90 & 95.00 & 81.90 \\

\bottomrule
\end{tabular}%
}
\vspace{-12pt}
\end{table}

We demonstrate that incorporating \textbf{Multiple-Choice QA} can effectively enhance a model’s capability on fundamental tasks. 
Tab.~\ref{tab:ab1} illustrates the impact of introducing Multiple-Choice QA on the performance of \textbf{Binary \& Type QA}.

From top to bottom, the rows correspond to: 
(1) the capability of the base model; 
(2) the performance obtained by applying RL on the base model \emph{without} using multiple-choice questions; 
(3) the performance of first performing SFT and then RL, while \emph{not} using multiple-choice questions in either stage; 
(4) the performance of first performing SFT and then RL, where multiple-choice questions are introduced only in the RL stage; 
and finally, 
(5) the performance where multiple-choice questions are used in both stages.

We observe that Multiple-Choice QA effectively transforms open-ended content descriptions into fixed-choice questions, and that training with such questions substantially improves the model’s perceptual capability. All experiments are conducted under identical parameter settings and with the same number of training steps.

\begin{table}[h]
\centering
\caption{The following shows the accuracy of models at different stages of multi-stage training on Multiple-Choice QA (MCQ) and Binary \& Type QA. For Binary \& Type QA, the reported values are the accuracy of samples where both the answer and the problem type are predicted correctly.
}
\vspace{2pt}
\label{tab:ab2}
\scriptsize
\setlength{\tabcolsep}{5pt}
\renewcommand{\arraystretch}{1.12}

\resizebox{\columnwidth}{!}{%
\begin{tabular}{
l
@{\hspace{3pt}\vrule width 0.4pt\hspace{3pt}} 
>{\centering\arraybackslash}c                  
@{\hspace{3pt}\vrule width 0.4pt\hspace{3pt}} 
*{3}{>{\centering\arraybackslash}c}            
@{\hspace{3pt}\vrule width 0.4pt\hspace{3pt}} 
>{\centering\arraybackslash}c                  
}
\toprule
\multirow{2}{*}{\textbf{Models}}
& \multicolumn{1}{c@{\hspace{3pt}\vrule width 0.4pt\hspace{3pt}}}{\textbf{MCQ}}
& \multicolumn{3}{c@{\hspace{3pt}\vrule width 0.4pt\hspace{3pt}}}{\textbf{Binary \& Type QA}}
& \multirow{2}{*}{\textbf{Total}} \\
\cmidrule(lr){2-2}\cmidrule(lr){3-5}
& Avg. & Structure & Semantic & Low-level  & \\
\midrule

Base & 41.90 & 53.50 & 21.40 & 24.60 & 37.53 \\  
StableI2I-Dev.1 & 92.37 & 83.70 & 59.00 & 93.70 & 85.58 \\
StableI2I-Dev.2 & 91.17 & 80.30 & 67.70 & 96.60 & 86.35 \\
StableI2I & 91.70 & 83.70 & 67.30 & 98.00 & 87.35 \\

\bottomrule
\end{tabular}%
}
\vspace{-20pt}
\end{table}

Tab.~\ref{tab:ab2} shows the effect of our multi-stage training and data augmentation strategy on model performance, and the overall training pipeline is illustrated in Fig.~\ref{fig:training_pipeline}.
StableI2I-Dev.1 denotes the first-stage SFT model. Owing to the limited amount of editing data, it performs poorly on the \textbf{Semantic} dimension.
After RL training, StableI2I-Dev.2 significantly improves the accuracy on this dimension; however, the reduced diversity of RL data leads to performance degradation on several other categories.
By combining the augmented data used in StableI2I-Dev.2 with the first-stage data and re-training the base model via SFT, we obtain the final model StableI2I.
Compared with StableI2I-Dev.1, StableI2I achieves substantially better overall performance, while largely retaining the gains of StableI2I-Dev.2.
These results confirm the effectiveness of our data augmentation strategy in improving overall model performance.

\vspace{-7pt}

\section{Conclusion}
\vspace{-5pt}
StableI2I is the first framework to systematically evaluate fidelity in image-to-image (I2I) tasks from both semantic and pixel-level perspectives. It enables reliable assessment of whether generative models preserve critical visual information and provides StableI2I-Bench as a precise benchmark for evaluating MLLMs under multi-image consistency constraints. By requiring consistency across multiple images at both semantic and pixel levels, this benchmark poses a substantial challenge to existing MLLMs and serves as an effective tool for measuring perceptual ability beyond single-image understanding. We believe that the introduction of StableI2I can substantially improve the generation quality of I2I models, leading to more faithful, realistic, and perceptually consistent outputs.

\textbf{Limitations.} For detailed subjective visualizations and analysis of specific failure cases, please refer to Appendix~\ref{sec:B3}.

\section*{Impact Statement}
This paper introduces StableI2I, a unified evaluation framework designed to enhance the reliability of image-to-image (I2I) transitions by detecting unintended semantic and structural drift. By establishing a principled methodology for measuring content fidelity across semantic, structural, and low-level appearance dimensions without requiring reference images, our work provides a critical diagnostic tool for high-stakes applications such as medical imaging and remote sensing, where information consistency is paramount. While StableI2I facilitates the development of more trustworthy generative systems, we acknowledge that automated fidelity assessments must be continually audited for potential biases in "consistency" definitions to ensure the framework remains inclusive of diverse visual domains and cultural contexts.

\bibliography{example_paper}
\bibliographystyle{icml2026}

\newpage
\appendix
\onecolumn
\section{Dataset}
\label{sec:A}
This section mainly provides supplementary details on the data sources, the models used for data construction, the human annotation workflow, and the prompts adopted in the data generation process.

\subsection{Data Construction and Statistics}
\label{sec:A1}
Our data pipeline primarily leverages images from ImageNet~\cite{Imagenet}, Unsplash~\cite{unsplash}, DIV2K~\cite{div2k}, Alchemist~\cite{alchemist} and ArtiMuse~\cite{artimuse}. The exact number of images drawn from each source is reported in Tab.~\ref{tab:app_data_stats}. After collection, we first constrain the maximum side length of each image to be no greater than 1344 pixels; images exceeding this limit are resized with preserved aspect ratio.

For image restoration tasks, we apply the ESRGAN~\cite{esrgan} degradation pipeline to synthesize degraded inputs. We then perform text-guided image restoration using three models: SeeSR~\cite{Seesr}, SUPIR~\cite{supir}, and OSEDiff~\cite{osediff}. To further increase the diversity of generated samples, a subset of the restored images is subjected to a second-stage enhancement, mainly using SwinIR~\cite{swinir} and ESRGAN~\cite{esrgan}.
For restoration results obtained under real-world settings, we use PromptIR~\cite{promptir}, OSEDiff~\cite{osediff}, Qwen-Image-Edit-2509~\cite{Qwen-Image}, and Bagel~\cite{bagel}. Specifically, we constructed our Low-level Dataset by randomly sampling 500 images from several public datasets and applying the ESRGAN degradation pipeline, as detailed in Tab.~\ref{tab:low_level_dataset_construction}.

\begin{table}[h]
\centering
\caption{Constuction of the Low-level Dataset.}
\label{tab:low_level_dataset_construction}
\begin{tabular}{@{}llc@{}}
\toprule
\textbf{Type} & \textbf{Source Dataset} & \textbf{\# Sample} \\ \midrule
Low-Light Enhancement & LOL~\cite{lol} & 409 \\
Image Dehazing & Nature20~\cite{nature20} & 190 \\
Image Dehazing & O-Haze~\cite{O-HAZE_2018} & 40 \\
Deraining & Rain800~\cite{rain800} & 661 \\
Raindrop Removal & RainDrop~\cite{raindrop} & 861 \\
Underwater Image Enhancement & UIEB~\cite{uieb} & 700 \\
Image Super-Resolution & DIV2K~\cite{div2k} & 800 \\ \midrule
\textbf{Total} & & \textbf{3661} \\ \bottomrule
\end{tabular}
\end{table}

For image editing tasks, the data are generated using Qwen-Image-Edit-2509~\cite{Qwen-Image}, OmniGen~\cite{omnigen}, SD3~\cite{sd3}, and GPT-Image-1~\cite{openai_gpt_image}. 
Our editing instructions are primarily constructed by prompting GPT-5 to integrate the visual information in the image and to formulate edits based on three major categories—add, replace, and remove—resulting in a concise editing instruction.
The number of samples constructed in the first stage is also reported in Tab.~\ref{tab:app_data_stats}.

\begin{table}[h]
\centering
\caption{Data usage and annotation statistics. \emph{Total} denotes the number of intact images after download and removal of corrupted files. \emph{Image Restoration} and \emph{Image Editing} indicate the numbers of successfully synthesized samples produced by their respective pipelines.
\textbf{In \emph{Image Restoration}}, \emph{GPT-5} refers to samples initially annotated using GPT-5, selected from the synthesized data and successfully labeled. \emph{Human} denotes the subset randomly sampled from the GPT-5--annotated data (15{,}000 samples) and subsequently cleaned and verified by human annotators.
\textbf{In \emph{Image Editing}}, \emph{V1-Human} denotes the number of samples annotated by human annotators in the first stage, and \emph{V2-Enhance} denotes the number of samples annotated using StableI2I-Dev.2 after multi-stage training.}
\label{tab:app_data_stats}
\setlength{\tabcolsep}{2pt}
\renewcommand{\arraystretch}{1.1}
\begin{tabular}{
l
l @{\hspace{4pt}\vrule width 0.8pt\hspace{4pt}}
>{\columncolor{gray!12}}c
c
c @{\hspace{4pt}\vrule width 0.8pt\hspace{4pt}}
>{\columncolor{gray!12}}c
c
c
}
\toprule
\textbf{Source} & \textbf{Total} & \multicolumn{1}{c}{\textbf{Image Restoration}} & GPT-5 & Human & \multicolumn{1}{c}{\textbf{Image Editing}} & V1-Human & V2-Enhance \\
\midrule
ArtiMuse   & 10002  & 4587  & \multirow{5}{*}{70640} & \multirow{5}{*}{6722} & 9933  & --   & 9933  \\
ImageNet   & 136590 & 89490 &                         &                         &  --     &  --    &  --     \\
DIV2K      & 800    & 800   &                         &                         & 800   & 800  & --    \\
Unsplash   & 24996  & 19264 &                         &                         & 24996 & --   & 24996 \\
Alchemist  & 4039   & 4039  &                         &                         & 4039  & 4039 & --    \\
\bottomrule
\end{tabular}
\end{table}

\subsection{Human Annotation and Annotator Training}
\label{sec:A2}
We first issued a public tender for the annotation task, and three professional annotation companies submitted bids. For evaluation, we selected 100 samples for each task and provided detailed annotation guidelines, asking all three companies to conduct a pilot annotation. The pilot annotation accuracies achieved by the three companies were 0.852, 0.725, and 0.700, respectively. We selected the company with the highest accuracy to carry out the full-scale annotation.

The final dataset consists of two parts: image restoration data (15,000 samples randomly drawn from GPT-5 coarse annotations) and image editing data (4,839 samples). Each task was annotated by a team of 10 annotators over seven consecutive working days, using a cross-annotation protocol that included one round of annotation followed by a review phase. Prior to annotation, all annotators received video-based training. During the annotation process, questions were addressed in real time through a shared document.

For the image restoration data, annotators were only required to judge whether GPT-5’s answer was correct and to label the level of degradation using a three-point scale. For the image editing data, annotators were required to assign an error type, optionally use bounding boxes to localize the problematic regions when necessary, and finally label the level of degradation using the same three-point scale.

In total, we obtained 6,722 human-annotated image restoration samples and 4,839 image editing samples. At acceptance, the overall annotation pass accuracy reached 94\%.

\subsection{Detailed Description of StableI2I-Bench}
\label{sec:A3}
All data in StableI2I-Bench are based on human-annotated samples, and the dataset has no overlap with the data used in Stable-Train.

We first present the StableI2I-Bench evaluation templates, followed by representative examples from the benchmark.

\subsubsection{StableI2I-Bench evaluation templates}

We adopt a \textbf{fixed prompt template} for all evaluation samples. Each prompt consists of four parts:
(1) two input images,
(2) the corresponding I2I instruction (prompt),
(3) a description of the prerequisite knowledge, and
(4) a specification of the required output format.
The benchmark covers three evaluation dimensions: \textbf{Semantic Level}, \textbf{Structure Level}, and \textbf{Low-level Appearance}. Each dimension contains 1000 image pairs, and within each dimension, the proportion of samples labeled as ``no issue'' does not exceed 50\%.
Below we provide detailed descriptions of the prompt design for the three evaluation dimensions.

\newtcolorbox{promptbox}[1]{%
  enhanced jigsaw,
  breakable,
  colback=white,
  colframe=black,
  boxrule=0.8pt,
  arc=3mm,
  outer arc=3mm,
  left=6pt,right=6pt,top=6pt,bottom=6pt,
  fonttitle=\bfseries\small,
  coltitle=white,
  colbacktitle=black,
  boxed title style={
    arc=3mm,
    outer arc=3mm,
    boxrule=0pt,
    interior style={fill=black},
  },
  title={#1},
  pad at break*=2mm,
  before skip=6pt,
  after skip=6pt,
}

\begin{promptbox}{Semantic Level}
\small
\textbf{The first image } \texttt{<image>}: Before processing.\\
\textbf{The second image } \texttt{<image>}: After processing.\\[2pt]

\textbf{The task prompt is:} \texttt{<TASK\_PROMPT>}\\[2pt]

\textbf{Task Guidelines:}\\[-1pt]
{\ttfamily\small
Please evaluate this image-to-image (I2I) transition from a semantic content fidelity perspective.

Compare the output image strictly against the input image, conditioned on the given task prompt. Your goal is to determine whether any regions that should remain unchanged have undergone unintended semantic changes.

Specifically, check whether the output image introduces any of the following semantic inconsistencies relative to the input image:

ADD: new objects, parts, text, symbols, or meaningful elements appear that are not implied by the task prompt.

REMOVE: existing objects, parts, text, or meaningful elements in the input image are missing in the output image.

REPLACE: an existing object, part, or attribute is substituted with a different semantic entity (e.g., a dog becomes a cat), or a meaningful attribute changes (e.g., ``red'' becomes ``blue'') when such change is not implied by the task prompt.

IMPORTANT GUIDELINES:
1) Focus on semantic content only. Ignore purely low-level appearance differences (e.g., mild noise or compression artifacts) unless they cause an actual semantic change (e.g., text becomes unreadable).
2) Legitimate global side effects that are a physically plausible consequence of the intended edit (e.g., shadows, reflections, or minor lighting changes) should not be counted as semantic errors.
3) If the task prompt is NULL (no specified edit or restoration intent), then the expected behavior is identity mapping: the two images should be completely identical in semantic content. Any semantic difference should be marked as inconsistent.
4) Use "No" whenever you detect any potential semantic inconsistency in regions that should have been preserved.}\par\vspace{2pt}

\textbf{Output Format:}\\[-1pt]
{\ttfamily\small
Return your decision in a single line of valid JSON with the format:
\{"answer": "Yes", "problem": "NULL"\} if the images are semantically consistent,
otherwise \{"answer": "No", "problem": ["add", "replace", "remove"]\}.}

\end{promptbox}

\begin{promptbox}{Structure Level}
\small
\textbf{The first image } \texttt{<image>}: Before processing.\\
\textbf{The second image } \texttt{<image>}: After processing.\\[2pt]

\textbf{The task prompt is:} \texttt{<TASK\_PROMPT>}\\[2pt]

\textbf{Task Guidelines:}\\[-1pt]
{\ttfamily\small
Please evaluate this image-to-image (I2I) transition from a structural
and texture consistency perspective.

Compare the output image strictly against the input image, conditioned
on the given task prompt. Your goal is to determine whether any regions
that are expected to remain unchanged have undergone unintended
structural or texture-level changes.

Specifically, check for the following types of inconsistencies:

Misalignment: global or local geometric distortions, spatial shifts,
  shape warping, incorrect object boundaries, layout drift, or broken
  structural coherence relative to the input image.

Repainting: unintended re-rendering of textures, materials, fine
  surface details, or local appearance patterns in regions that should
  have been preserved (e.g., skin, background, clothing, walls, ground,
  hair texture outside the edited region).

Important guidelines:
1) Only judge regions that are NOT explicitly targeted by the task
   prompt. Any change that is a necessary and physically plausible
   consequence of the intended edit (e.g., lighting, shading, subtle
   color adaptation) should NOT be counted as an error.
2) Focus on structure and texture consistency only. Ignore purely
   semantic category changes unless they manifest as clear repainting
   or structural deformation.
3) If the task prompt is NULL (no specified edit/restoration intent),
   then the expected behavior is identity mapping: the two images should
   be completely identical in structure and texture. Any deviation
   should be marked as inconsistent.
4) If both misalignment and repainting are observed, list both.
5) When uncertain, choose ``No''(i.e., favor sensitivity over specificity).}\par\vspace{2pt}

\textbf{Output Format:}\\[-1pt]
{\ttfamily\small
Return your decision in a single line of valid JSON with the format:
\{"answer": "Yes", "problem": "NULL"\} if the images are consistent,
otherwise \{"answer": "No", "problem": ["misalignment", "repainting"]\},
where the "problem" field should reflect the dominant issue(s) observed.}
\end{promptbox}

\begin{promptbox}{Low-level Appearance}
\small
\textbf{The first image } \texttt{<image>}: Before processing.\\
\textbf{The second image } \texttt{<image>}: After processing.\\[2pt]

\textbf{The task prompt is:} \texttt{<TASK\_PROMPT>}\\[2pt]

\textbf{Task Guidelines:}\\[-1pt]
{\ttfamily\small
Please evaluate this image-to-image (I2I) transition from a low-level
visual fidelity perspective.

Compare the output image strictly against the input image and determine
whether the processing has introduced any unintended low-level visual
degradation or distributional shift.

Specifically, check for the presence of any of the following issues in
the output image relative to the input image:

Blur: loss of sharpness or edge detail,

Noise: newly introduced random pixel-level noise or grain,

Color cast: unintended global or local color shifts,

Exposure degradation: over-exposure, under-exposure, or brightness/contrast distortion,

Artifact: compression artifacts, ringing, blocking, haloing, or other synthetic patterns.

This task focuses only on unintended low-level changes. Do NOT consider
high-level semantic differences or structural changes.

If the processing described in the task prompt is explicitly intended to
produce any of the above effects (e.g., denoising, deblurring, color
correction, artifact removal, exposure adjustment), then this case
should be ignored.

If no specific task type is given (i.e., the task prompt is NULL),
simply judge whether the two images are pixel-wise and perceptually
identical, up to negligible numerical or compression differences.

}\par\vspace{2pt}

\textbf{Output Format:}\\[-1pt]
{\ttfamily\small
Return your decision in a single line of valid JSON with the format:
In the ``ignored" case, output:
\{"answer": "NULL", "problem": "NULL"\}
If the output image is consistent with the input image at the low-level
appearance:
\{"answer": "Yes", "problem": "NULL"\}
Otherwise, if any unintended low-level degradation or shift is detected:
\{"answer": "No", "problem": ["noise", "blur", "color cast", "exposure degradation", "artifact"]\}}
\end{promptbox}

\subsubsection{StableI2I-Bench Examples}
Below, we randomly select three benchmark entries for illustration, corresponding to examples at the Structure level, Semantic level, and Low-level Appearance level, respectively.

\begin{promptbox}{StableI2I-Bench: Structure Level I2I Evaluation Example}
\small
\textbf{ID:} \texttt{test\_30}\\[2pt]

\textbf{Images:}\\[-2pt]
{\ttfamily\small
Input image: \texttt{/XXX/alchemist/image/9900644e50f1a52f5d5bb172e5bda971.jpg}\\
Output image: \texttt{/XXX/alchemist/image\_edit/9900644e50f1a52f5d5bb172e5bda971.jpg}
}\\[4pt]

\textbf{Human Prompt:}\\[-1pt]
{\ttfamily\small
The first image <image>: Before processing.\\
The second image <image>: After processing.\\
The task prompt is: Replace the gray jacket with a light pastel-colored cardigan to complement the spring setting.\\
Please evaluate this image-to-image (I2I) transition from a structural\\
and texture consistency perspective.\\

Compare the output image strictly against the input image, conditioned\\
on the given task prompt. Your goal is to determine whether any regions\\
that are expected to remain unchanged have undergone unintended\\
structural or texture-level changes.\\

Specifically, check for the following types of inconsistencies:\\

- misalignment: global or local geometric distortions, spatial shifts,\\
  shape warping, incorrect object boundaries, layout drift, or broken\\
  structural coherence relative to the input image.\\

- repainting: unintended re-rendering of textures, materials, fine\\
  surface details, or local appearance patterns in regions that should\\
  have been preserved (e.g., skin, background, clothing, walls, ground,\\
  hair texture outside the edited region).\\

Important guidelines:\\
1) Only judge regions that are NOT explicitly targeted by the task\\
   prompt. Any change that is a necessary and physically plausible\\
   consequence of the intended edit (e.g., lighting, shading, subtle\\
   color adaptation) should NOT be counted as an error.\\
2) Focus on structure and texture consistency only. Ignore purely\\
   semantic category changes unless they manifest as clear repainting\\
   or structural deformation.\\
3) If the task prompt is NULL (no specified edit/restoration intent),\\
   then the expected behavior is identity mapping: the two images should\\
   be completely identical in structure and texture. Any deviation\\
   should be marked as inconsistent.\\
4) If both misalignment and repainting are observed, list both.\\
5) When uncertain, choose ``No'' (i.e., favor sensitivity over specificity).\\

Return your decision in a single line of valid JSON with the format:\\
\{"answer": "Yes", "problem": "NULL"\} if the images are consistent,\\
otherwise \{"answer": "No", "problem": ["misalignment", "repainting"]\},\\
where the "problem" field should reflect the dominant issue(s) observed.
}\\[4pt]

\textbf{Model Output (GT):}\\[-1pt]
{\ttfamily\small
\{"answer": "No", "problem": ["repainting"]\}
}

\end{promptbox}

\begin{promptbox}{StableI2I-Bench: Semantic Level I2I Evaluation Example}
\small
\textbf{ID:} \texttt{test\_1589}\\[2pt]

\textbf{Images:}\\[-2pt]
{\ttfamily\small
Input image: \texttt{/XXX/imagenet/lq\_input/n02396427\_11580.jpg}\\
Output image: \texttt{/XXX/imagenet/imagenet100\_new\_x8\_out2\_select\_15\_combine/n02396427\_11580.jpg}
}\\[4pt]

\textbf{Human Prompt:}\\[-1pt]
{\ttfamily\small
The first image <image>: Before processing.\\
The second image <image>: After processing.\\
The task prompt is: Image restoration.\\
Please evaluate this image-to-image (I2I) transition from a semantic\\
content fidelity perspective.\\

Compare the output image strictly against the input image, conditioned\\
on the given task prompt. Your goal is to determine whether any regions\\
that should remain unchanged have undergone unintended semantic changes.\\

Specifically, check whether the output image introduces any of the\\
following semantic inconsistencies relative to the input image:\\
- add: new objects, parts, text, symbols, or meaningful elements appear\\
       that are not implied by the task prompt.\\
- remove: existing objects, parts, text, or meaningful elements in the\\
          input image are missing in the output image.\\
- replace: an existing object/part/attribute is substituted with a\\
           different semantic entity (e.g., a dog becomes a cat), or a\\
           meaningful attribute changes (e.g., ``red'' becomes ``blue'')\\
           when such change is not implied by the task prompt.\\

Important guidelines:\\
1) Focus on semantic content only. Ignore purely low-level appearance\\
   differences (e.g., mild noise, compression artifacts) unless they\\
   cause an actual semantic change (e.g., text becomes unreadable).\\
2) Legitimate global side effects that are a physically plausible\\
   consequence of the intended edit (e.g., shadows, reflections, minor\\
   lighting changes) should NOT be counted as semantic errors.\\
3) If the task prompt is NULL (no specified edit/restoration intent),\\
   then the expected behavior is identity mapping: the two images should\\
   be completely identical in semantic content. Any semantic difference\\
   should be marked as inconsistent.\\
4) Use ``No'' whenever you detect any potential semantic inconsistency in\\
   regions that should have been preserved.\\

Return your decision in a single line of valid JSON with the format:\\
\{"answer": "Yes", "problem": "NULL"\} if the images are semantically consistent,\\
otherwise \{"answer": "No", "problem": ["add", "replace", "remove"]\}.
}\\[4pt]

\textbf{Model Output (GT):}\\[-1pt]
{\ttfamily\small
\{"answer": "No", "problem": ["replace"]\}
}

\end{promptbox}

\begin{promptbox}{StableI2I-Bench: Low-level Appearance I2I Evaluation Example}
\small
\textbf{ID:} \texttt{test\_2922}\\[2pt]

\textbf{Images:}\\[-2pt]
{\ttfamily\small
Input image: \texttt{/xxx/Low-level/ArtiMuse/pre/0\_158.jpg}\\
Output image: \texttt{/xxx/Low-level/ArtiMuse/post/0\_158.jpg}
}\\[4pt]

\textbf{Human Prompt:}\\[-1pt]
{\ttfamily\small
The first image <image>: Before processing.\\
The second image <image>: After processing.\\
The task prompt is: Modify the outfit to have a matte finish instead of a glossy one for a more subdued appearance..\\
Please evaluate this image-to-image (I2I) transition from a low-level\\
visual fidelity perspective.\\

Compare the output image strictly against the input image and determine\\
whether the processing has introduced any unintended low-level visual\\
degradation or distributional shift.\\

Specifically, check for the presence of any of the following issues in\\
the output image relative to the input image:\\
- blur: loss of sharpness or edge detail,\\
- noise: newly introduced random pixel-level noise or grain,\\
- color cast: unintended global or local color shifts,\\
- exposure degradation: over-exposure, under-exposure, or brightness/contrast distortion,\\
- artifact: compression artifacts, ringing, blocking, haloing, or other synthetic patterns.\\

This task focuses only on unintended low-level changes. Do NOT consider\\
high-level semantic differences or structural changes.\\

If the processing described in the task prompt is explicitly intended to\\
produce any of the above effects (e.g., denoising, deblurring, color\\
correction, artifact removal, exposure adjustment), then this case\\
should be ignored.\\

If no specific task type is given (i.e., the task prompt is NULL),\\
simply judge whether the two images are pixel-wise and perceptually\\
identical, up to negligible numerical or compression differences.\\

Return your decision in a single line of valid JSON with the format:\\

In the ``ignored'' case, output:\\
\{"answer": "NULL", "problem": "NULL"\}\\

If the output image is consistent with the input image at the low-level\\
appearance:\\
\{"answer": "Yes", "problem": "NULL"\}\\

Otherwise, if any unintended low-level degradation or shift is detected:\\
\{"answer": "No", "problem": ["noise", "blur", "color cast", "exposure degradation", "artifact"]\}
}\\[4pt]

\textbf{Model Output (GT):}\\[-1pt]
{\ttfamily\small
\{"answer": "No", "problem": ["blur"]\}
}

\end{promptbox}

\subsection{Detailed Description of StableI2I-Train}
\label{sec:A4}
\subsubsection{Supplementary Details on Data Synthesis and Scale of StableI2I-Train}
\label{sec:A41}

Here, we first provide supplementary explanations regarding the \textbf{Multiple-Choice QA} component. Since the earlier annotation process already identifies whether an error occurs and specifies the corresponding error type, we have sufficient information to construct multiple-choice questions.

The multiple-choice questions are mainly divided into two categories. The first category is \textbf{``Type''}, which focuses on identifying the type of error. The second category is \textbf{``Subtype''}, which focuses on identifying the object or region where the error occurs.

The first category of questions is relatively straightforward to construct. We can generate multiple-choice questions by enumerating all error types associated with the given category, together with an additional option indicating that none of the above answers is correct. Based on the error types involved in each sample, we then construct a multiple-selection question with a variable number of correct choices.

In contrast, the second category requires the model to understand which objects or regions are present in the image. Based on this understanding, we construct multiple-choice questions that include both objects that are incorrectly affected and objects that appear in the image but remain unaffected. This process requires the model to perform image understanding and question generation jointly.

The following two JSON examples illustrate the first and second categories of multiple-choice questions, respectively.

\begin{promptbox}{Multiple-Choice QA Example (Type)}
\small
\textbf{ID:} \texttt{test\_3674}\\[2pt]

\textbf{Images:}\\[-2pt]
{\ttfamily\small
Input image: \texttt{/XXX/DIV2K/DIV2K\_train\_HR/0781.png}\\
Output image: \texttt{/XXX/DIV2K/DIV2K\_train\_HR\_edit/0781.png}
}\\[4pt]

\textbf{Human Prompt:}\\[-1pt]
{\ttfamily\small
The first image <image>: Before processing.\\
The second image <image>: After processing.\\

Pick one answer. Given the request to image edit and Edit instruction is
Add a pair of aviator sunglasses to the giraffe's face, which types of
unintended semantic issues occurred in unchanged regions?\\

A. misalignment\\
B. No error observed in areas expected to remain unchanged\\
C. repainting\\

Return exactly ONE line of JSON with the format:\\
\{"answer":["A","B",...]\}\\
Examples:\\
Single-choice -> \{"answer":["D"]\}\\
Multi-choice  -> \{"answer":["A","C"]\}\\

Rules:\\
- "answer" MUST be a list even for single choice.\\
- Letters MUST be uppercase and chosen ONLY from the options shown.\\
- Output JSON only. No extra text.
}\\[4pt]

\textbf{Model Output (GT):}\\[-1pt]
{\ttfamily\small
\{"answer": ["C"]\}
}

\end{promptbox}

\begin{promptbox}{Multiple-Choice QA Example (Subtype)}
\small
\textbf{ID:} \texttt{test\_2053}\\[2pt]

\textbf{Images:}\\[-2pt]
{\ttfamily\small
Input image: \texttt{/XXX/imagenet/lq\_input/n02099849\_1606.jpg}\\
Output image: \texttt{/XXX/imagenet/imagenet100\_new\_x8\_out2\_select\_15\_combine/n02099849\_1606.jpg}
}\\[4pt]

\textbf{Human Prompt:}\\[-1pt]
{\ttfamily\small
The first image <image>: Before processing.\\
The second image <image>: After processing.\\

Pick one answer. Given the request to image restoration, what element was
replaced even though it should have stayed unchanged in areas expected
to remain unchanged?\\

A. Replacement involving dog → deer\\
B. Disappearance of background trees\\
C. Removal of grass\\
D. Shrubs missing\\

Return exactly ONE line of JSON with the format:\\
\{"answer":["A","B",...]\}\\
Examples:\\
Single-choice -> \{"answer":["D"]\}\\
Multi-choice  -> \{"answer":["A","C"]\}\\

Rules:\\
- "answer" MUST be a list even for single choice.\\
- Letters MUST be uppercase and chosen ONLY from the options shown.\\
- Output JSON only. No extra text.
}\\[4pt]

\textbf{Model Output (GT):}\\[-1pt]
{\ttfamily\small
\{"answer": ["A"]\}
}

\end{promptbox}

\begin{figure*}[h] 
\vspace{-5pt}
  \centering 
  \includegraphics[width=0.85\textwidth]{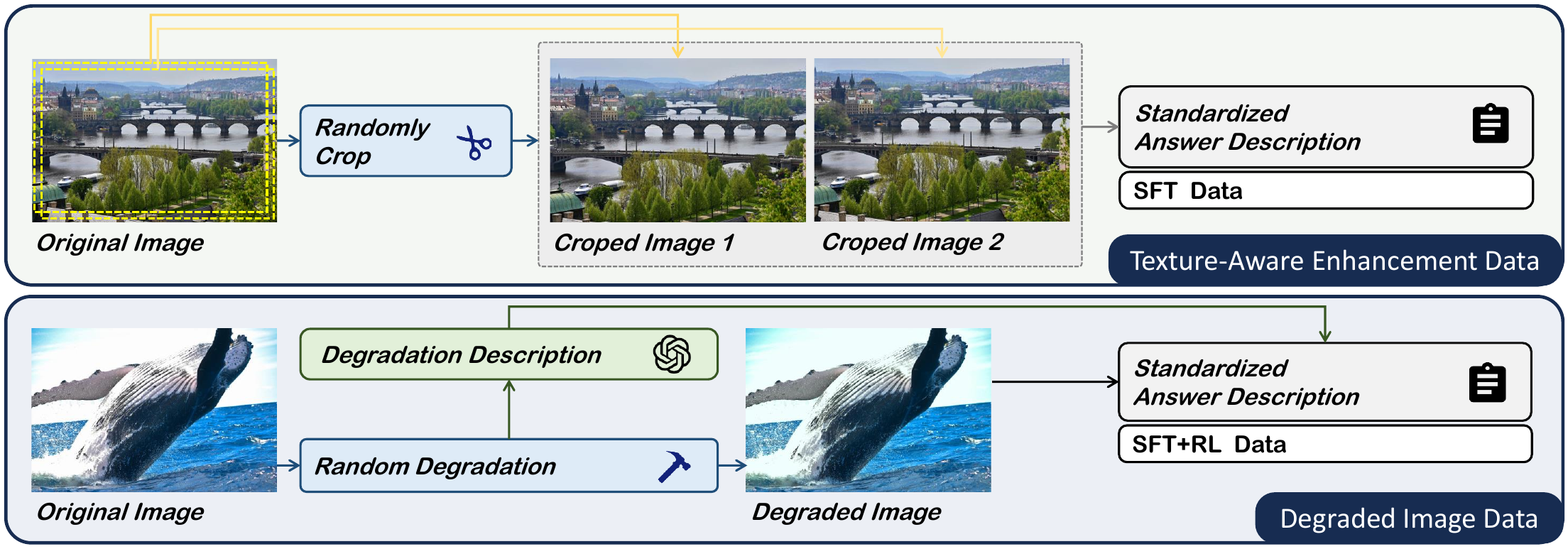} 
  \captionsetup{skip=1.5pt}
  \caption{An overview of the data construction pipelines for Texture-Aware Enhancement Data and Degraded Image Data.}
  \vspace{-5pt}
  \label{fig:data_app}
\end{figure*}

During training, we observed that existing multimodal large models exhibit weak pixel-level alignment capability. Therefore, in addition to the standard image restoration and image editing data, we introduced two auxiliary data types to enhance the perceptual capacity of the model encoder: \textbf{Texture-Aware Enhancement Data} and \textbf{Degraded Image Data}, as illustrated in Fig.~\ref{fig:data_app}.

Texture-Aware Enhancement Data are constructed by randomly cropping a natural image into two different images with a cropping ratio of 95–98\%. After this processing, the two images remain nearly identical in global semantic content, but the model is forced to attend to subtle pixel-level structural differences between them. This encourages the model to learn fine-grained pixel alignment and correspondence.
Degraded Image Data are designed to assist the model in learning degradation phenomena that arise in image-to-image (I2I) tasks. This subset covers multiple types of low-level appearance shifts, including blur, noise, compression artifacts, and color distortions.

\begin{table}[!h]
\centering
\caption{The training data volume for the \textbf{Binary \& Type QA} category is summarized as follows. 
Real denotes samples generated by the data synthesis pipeline with human annotations (Fig.~\ref{fig:data_pipeline}), 
while Synthetic refers to samples synthesized using the pipeline in Fig.~\ref{fig:data_app}. 
The latter subset is mainly introduced to enhance the model’s pixel-level perceptual capability.
 ``--'' denotes not applicable.}
\label{tab:data_stats_tasks1}
\setlength{\tabcolsep}{8pt}
\renewcommand{\arraystretch}{1.12}

\begin{tabular}{@{}l l r r r@{}}
\toprule
\textbf{Dimension} & \textbf{Source} & \textbf{Image Identity} & \textbf{Image Restoration} & \textbf{Image Editing} \\
\midrule

\multirow{2}{*}{Structure Level}
& Synthetic & 207{,}866 & 415{,}732 & -- \\
& Real      & --        & 6{,}722   & 38{,}904 \\
\midrule

Semantic Level
& Real      & --        & 70{,}640  & 54{,}447 \\
\midrule

Low-level Appearance
& Synthetic & 118{,}808 & 7{,}905   & 38{,}238 \\
\bottomrule
\end{tabular}
\end{table}

\begin{table}[!h]
\centering
\caption{The training data volume for the \textbf{Multiple-choice QA} category is reported as follows. 
\textbf{Type} refers to answering the type of error that occurs, while \textbf{Subtype} refers to answering the specific object or content involved in the error. 
All questions are formulated as multiple-answer multiple-choice questions. 
``--'' denotes not applicable.}
\label{tab:data_stats_tasks2}
\setlength{\tabcolsep}{8pt}
\renewcommand{\arraystretch}{1.15}

\begin{tabular}{@{}l l r r r@{}}
\toprule
\textbf{Dimension} & \textbf{QA Level} & \textbf{Image Identity} & \textbf{Image Restoration} & \textbf{Image Editing} \\
\midrule

Structure Level
& Type     & --        & 5{,}722 & 3{,}975 \\
\midrule

\multirow{2}{*}{Semantic Level}
& Type     & --        & 8{,}406 & 4{,}851 \\
& Subtype  & --        & 7{,}992 & 1{,}451 \\
\midrule

\multirow{2}{*}{Low-level Appearance}
& Type     & --        & 5{,}186 & 2{,}995 \\
& Subtype  & 2{,}000   & 2{,}575 & --      \\

\bottomrule
\end{tabular}
\end{table}

\begin{table}[!h]
\centering
\caption{The training data volume for the \textbf{Open-ended QA} category is summarized as follows. 
Synthetic denotes answer pairs synthesized from the annotated errors in the data construction pipeline and their corresponding fine-grained error descriptions, 
while GPT-5 refers to samples whose \textit{think} rationales and answers to fine-grained error questions are generated by GPT-5.
``--'' denotes not applicable.}
\label{tab:data_stats_tasks3}
\setlength{\tabcolsep}{10pt}
\renewcommand{\arraystretch}{1.15}

\begin{tabular}{@{}l l r r@{}}
\toprule
\textbf{Dimension} & \textbf{Source} & \textbf{Image Restoration} & \textbf{Image Editing} \\
\midrule

\multirow{2}{*}{Semantic Level}
& Synthetic & 70{,}640 & 4{,}851 \\
& GPT-5       & 8{,}406  & 34{,}929 \\
\midrule

Low-level Appearance
& Synthetic & 8{,}599  & 34{,}929 \\
\bottomrule

\end{tabular}
\end{table}

\begin{figure}[!h] 
\vspace{-5pt}
  \centering 
  \includegraphics[width=0.5\textwidth]{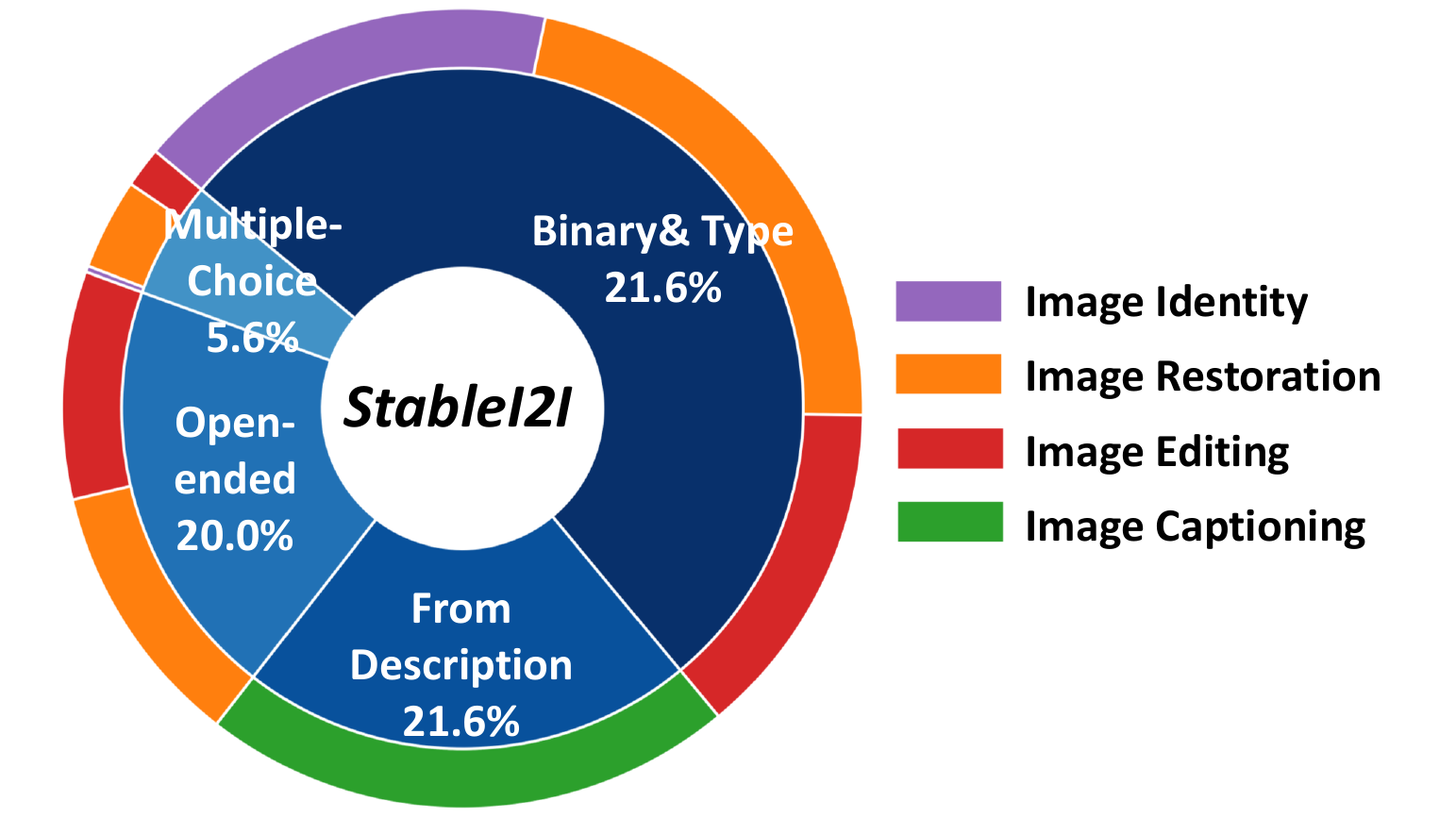} 
  \captionsetup{skip=1.5pt}
  \caption{The proportions of the four types of training data used for SFT.
The overall taxonomy of image-to-image (I2I) tasks is divided into three categories: Image Editing, Image Restoration, and Image Identity, where Image Identity refers to directly comparing two images without applying any transformation.}
  \vspace{-5pt}
  \label{fig:data_app_propotion}
\end{figure}

Building upon these data types, we further introduce an additional task, termed Image Identity, which requires the model to determine whether two identical images exhibit any differences. This task is designed to strengthen the model’s ability to perceive fine-grained pixel-level correspondences across multiple images. The images for this task are constructed from the COCO~\cite{coco} dataset following the same pipelines used for Texture-Aware Enhancement Data and Degraded Image Data.

The final data composition ratios at the SFT stage are summarized as follows. The number of Free-form Descriptive samples is 174,866. The statistics of Binary \& Type QA are reported in Tab.~\ref{tab:data_stats_tasks1}, those of Open-ended QA are reported in Tab.~\ref{tab:data_stats_tasks3}, and those of Multiple-choice QA are reported in Tab.~\ref{tab:data_stats_tasks2}. The overall proportion of the training data is illustrated in Fig.~\ref{fig:data_app_propotion}.

\subsubsection{Input Prompt Template of StableI2I-Train}
\label{sec:A42}
During training, we adopt a fixed prompt template to better adapt the model to the target tasks. 
\textbf{Binary \& Type QA (Format~\ref{fig:output-format})} covers three evaluation dimensions: Semantic Level, Structure Level, and Low-level Appearance. 
\textbf{Open-ended QA (Format~\ref{fig:detailed-output-format})} covers two dimensions: Semantic Level and Low-level Appearance, 
since the Structure Level corresponds to global changes and does not admit more fine-grained textual descriptions.

We next describe the prompt designs for the three dimensions in Format~\ref{fig:output-format}, 
followed by the two prompt designs for the dimensions in Format~\ref{fig:detailed-output-format}.

\begin{promptbox}{Format 1: Structure Level}
\small
\textbf{The first image } \texttt{<image>}: Before processing.\\
\textbf{The second image } \texttt{<image>}: After processing.\\[2pt]

\textbf{The task prompt is:} \texttt{<TASK\_PROMPT>}\\[2pt]

\textbf{Task Guidelines:}\\[-1pt]
{\ttfamily\small
Please determine whether the texture-consistent regions that should remain unchanged between the pre- and post-process images are indeed consistent (e.g., unchanged areas remain identical, or for restoration tasks, whether the overall texture and color are consistent).
If no specific task type is given, simply judge whether the two images are identical.
Major inconsistencies generally fall into two categories: structural misalignment, texture repainting.
}\par\vspace{2pt}

\textbf{Output Format:}\\[-1pt]
{\ttfamily\small
Return your decision in a single line of valid JSON with the format: \{"answer": "Yes", "problem": "NULL"\} if the images are consistent, otherwise \{"answer": "No", "problem": ["misalignment", "repainting"]\}, where the "problem" field should reflect the most dominant issue observed between the two images.}
\end{promptbox}

\begin{promptbox}{Format 1: Semantic Level}
\small
\textbf{The first image } \texttt{<image>}: Before processing.\\
\textbf{The second image } \texttt{<image>}: After processing.\\[2pt]

\textbf{The task prompt is:} \texttt{<TASK\_PROMPT>}\\[2pt]

\textbf{Task Guidelines:}\\[-1pt]
{\ttfamily\small
Please determine whether the regions that should remain unchanged between the pre- and post-processed images exhibit any semantic errors — that is, whether there are additions, deletions, or modifications of semantic content relative to the pre-processed image.
If no specific task type is provided, simply determine whether the two images are completely identical.
}\par\vspace{2pt}

\textbf{Output Format:}\\[-1pt]
{\ttfamily\small
Return your decision in a single line of valid JSON with the format:
\{"answer": "Yes", "problem": "NULL"\} if the images are consistent,
otherwise \{"answer": "No", "problem": ["add", "replace", "remove"]\}.
"No" is used whenever any potential inconsistency is detected.}
\end{promptbox}

\begin{promptbox}{Format 1: Low-level Appearance}
\small
\textbf{The first image } \texttt{<image>}: Before processing.\\
\textbf{The second image } \texttt{<image>}: After processing.\\[2pt]

\textbf{The task prompt is:} \texttt{<TASK\_PROMPT>}\\[2pt]

\textbf{Task Guidelines:}\\[-1pt]
{\ttfamily\small
Please determine whether the pre-processed image has undergone any low-level degradation or shift after processing. Specifically, check whether there is any degradation (e.g., blur, noise), color cast, or newly introduced artifacts. If the processing described in the task type is explicitly related to any of the above (e.g., denoising, deblurring, color correction, artifact removal), then this case should be ignored.
If no specific task type is given, simply judge whether the two images are identical.
}\par\vspace{2pt}

\textbf{Output Format:}\\[-1pt]
{\ttfamily\small
Return your decision in a single line of valid JSON with the format:
In the "ignored" case, output:\{"answer": "NULL", "problem": "NULL"\},
\{"answer": "Yes", "problem": "NULL"\} if the images are consistent,
otherwise \{"answer": "No", "problem": ["noise", "blur", "color cast", "exposure degradation", "artifact"]\}.}
\end{promptbox}

\begin{promptbox}{Format 2: Semantic Level}
\small
\textbf{The first image } \texttt{<image>}: Before processing.\\
\textbf{The second image } \texttt{<image>}: After processing.\\[2pt]

\textbf{The task prompt is:} \texttt{<TASK\_PROMPT>\\ Assume this example DOES contain issues: semantic drift has occurred within regions that should be preserved.
Drift type(s): <TYPES\_FROM\_FORMAT\_1\_RESULT>}\\[2pt]

\textbf{Task Guidelines:}\\[-1pt]
{\ttfamily\small
Your task is to analyze the problem strictly in two stages:

1) Preservation analysis (think):
   - Identify the intended edit target region(s) according to the task prompt.
   - Explicitly state which changes can be ignored because they fall inside the intended edit scope.
   - Identify the regions/elements that must be preserved (non-edit regions), and list them as a concrete checklist with brief justification.

2) Problem reporting (problem):
   - Report ONLY issues that violate the preservation analysis above.
   - If something was stated as ignorable or allowed-to-change in the think stage, it MUST NOT appear here.
   - Focus on preserved regions and explain the semantic drift clearly.
   - Use only the drift type keys that were provided above (Drift type(s): XXX).
}\par\vspace{2pt}

\textbf{Output Format:}\\[-1pt]
{\ttfamily\small
Output MUST be a single valid JSON object and nothing else:
\{
  "think": "Preservation analysis: intended edit target and ignorable changes first; then a checklist of preserved elements with justification.",
  "problem": \{
    "add": "Describe added content in preserved regions (if applicable).",
    "replace": "Describe replaced content in preserved regions (if applicable).",
    "remove": "Describe removed or missing content in preserved regions (if applicable)."
  \}
\}}
\end{promptbox}

\begin{promptbox}{Format 2: Low-level Appearance}
\small
\textbf{The first image } \texttt{<image>}: Before processing.\\
\textbf{The second image } \texttt{<image>}: After processing.\\[2pt]

\textbf{The task prompt is:} \texttt{<TASK\_PROMPT> \\You are evaluating whether the AFTER image introduces unintended LOW-LEVEL degradation/shift in regions that should be preserved. Candidate degradation type(s) (use ONLY these keys if applicable): <TYPES\_FROM\_FORMAT\_1\_RESULT>}\\[2pt]

\textbf{Task Guidelines:}\\[-1pt]
{\ttfamily\small
Your task is to analyze strictly in two stages:

1) Preservation \& scope analysis (think):
   - Identify the intended target region(s) implied by the task prompt.
   - State which changes are allowed ONLY if they occur strictly inside the intended target region(s).
   - Clarify that low-level degradations (noise/blur/color cast/exposure issues/artifacts) are NOT intended unless the task prompt explicitly requests low-level enhancement/removal.
   - Provide a checklist of what must be preserved (non-target regions/elements) with brief justification.
   - If the task prompt explicitly requests low-level processing (e.g., denoise/deblur/color correction/exposure enhancement/artifact removal), then low-level changes consistent with that request and confined to the intended scope may be treated as allowed; state this in "think".

2) Problem reporting (problem):
   - Report ONLY low-level degradations that violate the scope above (i.e., occur in preserved regions or exceed intended scope).
   - If no violation is found, output an empty object: problem = {}.
   - Use ONLY the keys provided in YYY. Do not invent new keys.
   - For each key you include, describe: where it appears, how it differs from BEFORE, and the visual symptom.
}\par\vspace{2pt}

\textbf{Output Format:}\\[-1pt]
{\ttfamily\small
Output MUST be a single valid JSON object and nothing else:
\{
  "think": "…",
  "problem": \{
    "noise": "…",
    "blur": "…",
    "color cast": "…",
    "exposure degradation": "…",
    "artifact": "…"
  \}
\}}
\end{promptbox}


\section{Supplementary Experimental Results}
\label{sec:B}
\subsection{Supplementary Results of StableI2I and MLLMs on StableI2I-Bench}
\label{sec:B1}
As discussed in the main paper, the input prompt template used by StableI2I at inference time is not identical to the template adopted in StableI2I-Bench for evaluating general MLLMs. 
This is because StableI2I is a model specifically trained for I2I fidelity assessment, and its training relies on a fixed and unified task template. 
In contrast, the benchmark template contains richer instructional priors to enable general-purpose MLLMs to correctly perform the evaluation task.

\begin{table*}[h]
\centering
\caption{Quantitative comparison of mainstream models on StableI2I-Bench.
Binary Accuracy measures answer correctness, while Strict Accuracy additionally requires correct error types.
Best and second-best results are highlighted in \TopOne{dark blue} and \TopTwo{light blue}, respectively.}
\label{tab:cmp_model_app}

\scriptsize
\setlength{\tabcolsep}{3.5pt}
\renewcommand{\arraystretch}{1.05}

\begin{tabular}{
l
@{\hspace{4pt}\vrule width 0.4pt\hspace{4pt}}
*{4}{>{\centering\arraybackslash}c}
@{\hspace{4pt}\vrule width 0.4pt\hspace{4pt}}
*{4}{>{\centering\arraybackslash}c}
}
\toprule
\multirow{2}{*}{\textbf{Models}}
& \multicolumn{4}{c@{\hspace{4pt}\vrule width 0.4pt\hspace{4pt}}}{\textbf{Binary Accuracy}}
& \multicolumn{4}{c}{\textbf{Strict Accuracy}} \\
\cmidrule(lr){2-5} \cmidrule(lr){6-9}
& Structure & Semantic & Low-level & Avg.
& Structure & Semantic & Low-level & Avg. \\
\midrule

\rowcolor{blue!10}
\multicolumn{9}{l}{\textbf{Open-Source Models}} \\

Qwen3VL-8B-Instruct  
& 49.80 & 48.60 & 79.60 & 59.33
& 37.00 & 22.20 & 52.30 & 37.17 \\

Qwen3VL-32B-Instruct 
& 53.70 & 51.60 & 87.90 & 64.40
& 40.30 & 24.20 & 58.50 & 41.00 \\

InternVL-3.5-8B      
& 38.40 & 46.90 & 59.10 & 48.13
& 28.20 & 6.20  & 14.60 & 16.33 \\

InternVL-3.5-38B     
& 41.30 & 46.70 & 75.50 & 54.50
& 24.60 & 18.60 & 40.30 & 27.83 \\

\midrule

\rowcolor{gray!12}
\multicolumn{9}{l}{\textbf{Proprietary Models}} \\

Grok-4.1               
& 57.70 & 58.90 & 73.70 & 63.43
& 51.80 & 39.10 & 40.30 & 43.73 \\

Claude-Sonnet-4.5      
& 65.00 & 70.40 & 88.70 & 74.70
& 61.60 & 53.10 & 71.20 & 61.97 \\

Claude-Sonnet-4.5-think
& 65.40 & 66.50 & 86.30 & 72.73
& 62.90 & 51.90 & 68.60 & 61.13 \\

Gemini-2.5-pro         
& 68.17 & 77.50 & 92.50 & 79.39
& 63.36 & 55.30 & 60.10 & 59.59 \\

Gemini-3-pro           
& \TopTwo{70.77} & 80.14 & 83.98 & 78.30
& \TopTwo{64.26} & 59.17 & 61.06 & 61.50 \\

GPT-4o                 
& 58.50 & 80.10 & 88.10 & 75.57
& 53.90 & 60.30 & \TopTwo{76.30} & \TopTwo{63.50} \\

GPT-5                  
& 66.20 & \TopTwo{82.50} & \TopTwo{95.20} & \TopTwo{81.30}
& 61.00 & \TopTwo{64.10} & 63.20 & 62.77 \\

\midrule

StableI2I              
& \TopOne{85.40} & \TopOne{82.80} & \TopOne{99.10} & \TopOne{89.10}
& \TopOne{83.70} & \TopOne{67.30} & \TopOne{98.00} & \TopOne{83.00} \\

\bottomrule
\end{tabular}

\end{table*}

To provide a more controlled and fair comparison, we further conduct an additional experiment in which MLLMs are evaluated using the same input template as StableI2I. 
This setting allows us to isolate the effect of the prompt template and to better assess the intrinsic performance gap between StableI2I and general MLLMs under identical prompting conditions. The comparative results are shown in Tab.~\ref{tab:cmp_model_app}.

We observe that, compared with Tab.~\ref{tab:cmp_model} in the main paper, the performance of MLLMs drops the most on the Semantic Level dimension when the additional instructional priors are removed. 
In contrast, the performance on the Structure Level remains largely unchanged and even exhibits a slight improvement. 
We attribute this behavior to the fact that the Structure Level is inherently difficult for current models to judge. 
As a result, the presence or absence of additional priors does not substantially help models achieve reliable performance on this dimension.

Importantly, even under this significantly less informative prompting condition, StableI2I consistently outperforms all general-purpose MLLMs across all three evaluation dimensions. 
This result indicates that the superior performance of StableI2I does not stem from a more favorable or information-rich prompt design, but rather from its task-specific training and its enhanced pixel-level perceptual and alignment capabilities.

Moreover, this controlled comparison confirms that the benchmark prompt template does not confer an unfair advantage to StableI2I. 
On the contrary, the instruction-rich template in StableI2I-Bench primarily serves to make the evaluation task feasible for general-purpose MLLMs, rather than to artificially inflate their performance. 
When this auxiliary instructional prior is removed, MLLMs exhibit substantial performance degradation, whereas StableI2I remains robust and maintains strong performance across all dimensions.

These findings collectively demonstrate that the performance gap between StableI2I and general MLLMs reflects an intrinsic capability difference, rather than a confounding effect introduced by prompt engineering. 
This further validates the necessity of a task-specialized fidelity assessment model and highlights the limitations of current general-purpose MLLMs in fine-grained I2I fidelity evaluation.

\subsection{Ablation study with ImgEdit-Judge}
\label{sec:app}

Since the two models adopt different evaluation dimensions, it is difficult to directly assess single-image results.
Therefore, we compare the outputs produced by two different models and evaluate whether human preference judgments are consistent with the corresponding evaluation results.
Specifically, ImgEdit-Judge is evaluated using the \emph{Physical \& Detail Integrity} dimension.
Experiments are conducted on two benchmarks, \textbf{ImgEdit-Bench} and \textbf{GEdit-Bench}.
For each benchmark, we randomly sample 50 comparison pairs from a candidate pool consisting of Qwen-Image-Edit-2511~\cite{Qwen-Image}, Nano-Banana~\cite{google_gemini_image}, GPT-Image-1~\cite{openai_gpt_image}, and Omnigen2~\cite{omnigen2}.
The final results are reported in Tab.~\ref{tab:imgedit_judge_ablation}, showing that our model achieves better performance than ImgEdit-Judge in fidelity assessment.

\begin{table}[h]
\centering
\caption{Comparison of fidelity assessment accuracy between ImgEdit-Judge and StableI2I on ImgEdit-Bench and GEdit-Bench.}
\label{tab:imgedit_judge_ablation}
\small
\setlength{\tabcolsep}{10pt}
\renewcommand{\arraystretch}{1.2}
\begin{tabular}{c c c}
\toprule
\textbf{Accuracy} & \textbf{ImgEdit-Bench} & \textbf{GEdit-Bench} \\
\midrule
ImgEdit-Judge & 30\% & 52\% \\
StableI2I & 80\% & 76\% \\
\bottomrule
\end{tabular}
\end{table}

\subsection{Supplementary Results of StableI2I in Real-World I2I Evaluation Settings}
\label{sec:B2}
\begin{figure*}[h] 
\vspace{-5pt}
  \centering 
  \includegraphics[width=1\textwidth]{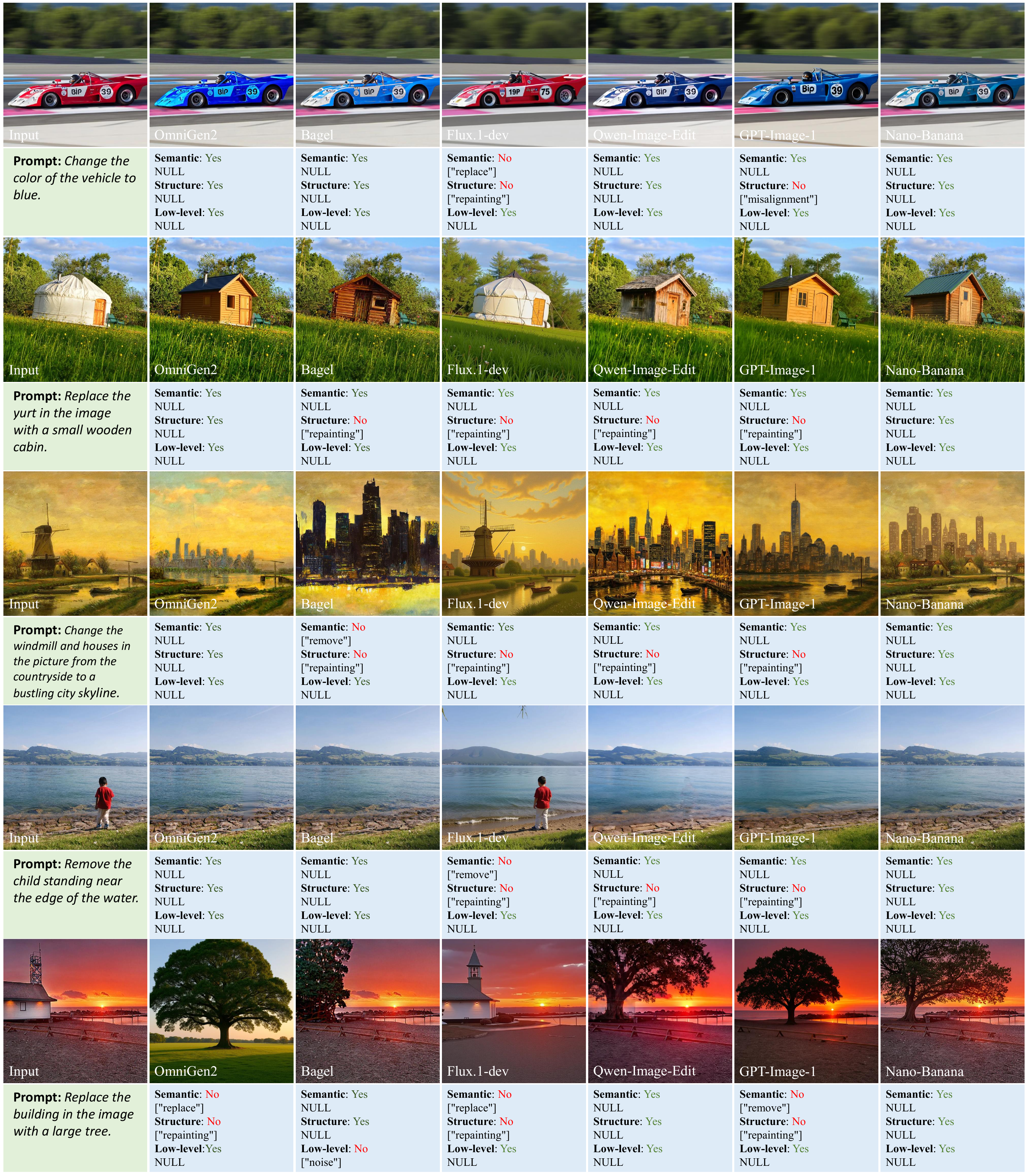} 
  \captionsetup{skip=1.5pt}
  \caption{Visualization of the evaluation results on ImgEdit-Bench using Format 1.}
  \vspace{-5pt}
  \label{fig:data_app_imgedit}
\end{figure*}

\begin{figure*}[h] 
\vspace{-5pt}
  \centering 
  \includegraphics[width=1\textwidth]{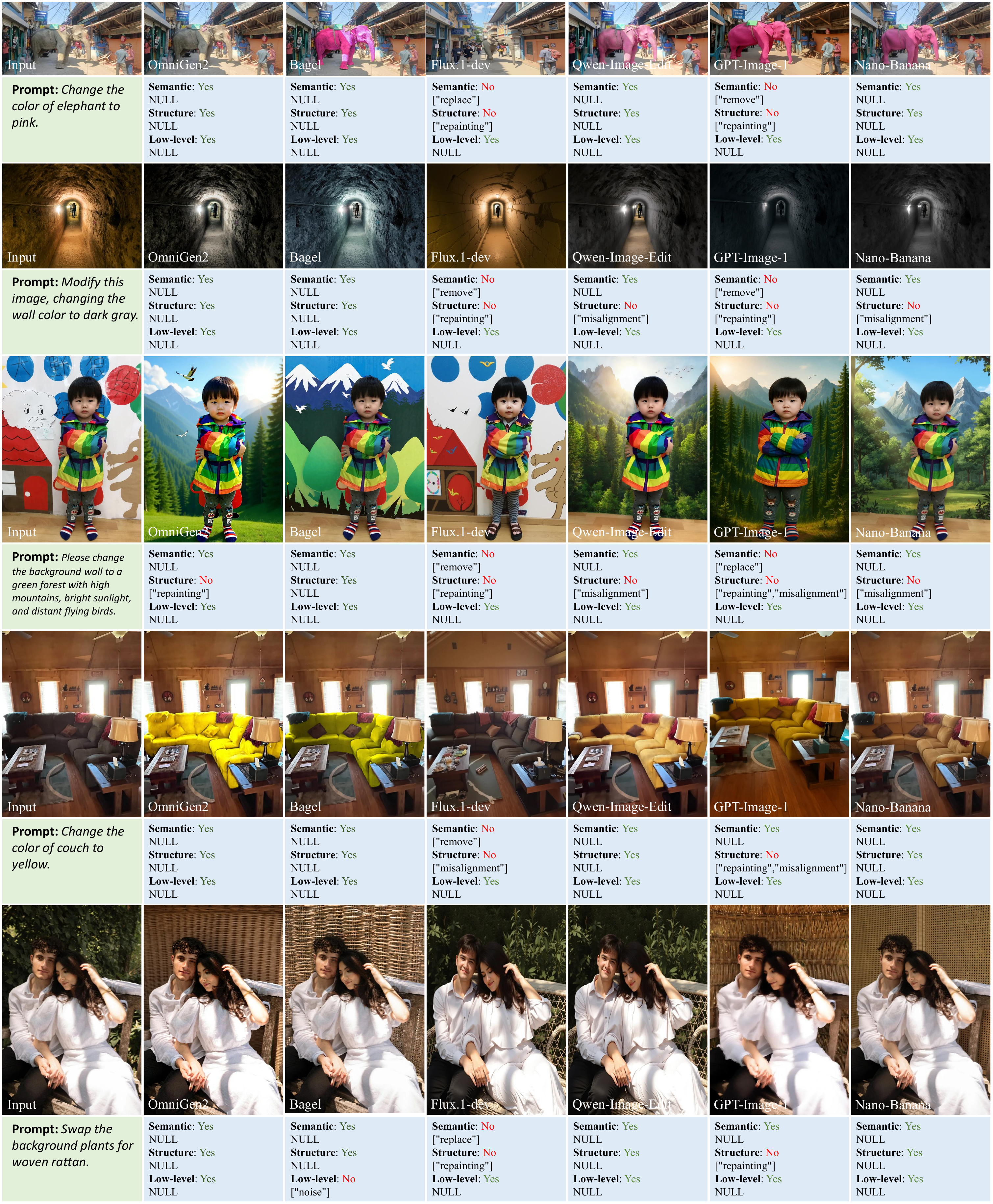} 
  \captionsetup{skip=1.5pt}
  \caption{Visualization of the evaluation results on GEdit-Bench using Format 1.}
  \vspace{-5pt}
  \label{fig:data_app_gedit}
\end{figure*}

\begin{figure*}[h] 
\vspace{-5pt}
  \centering 
  \includegraphics[width=1\textwidth]{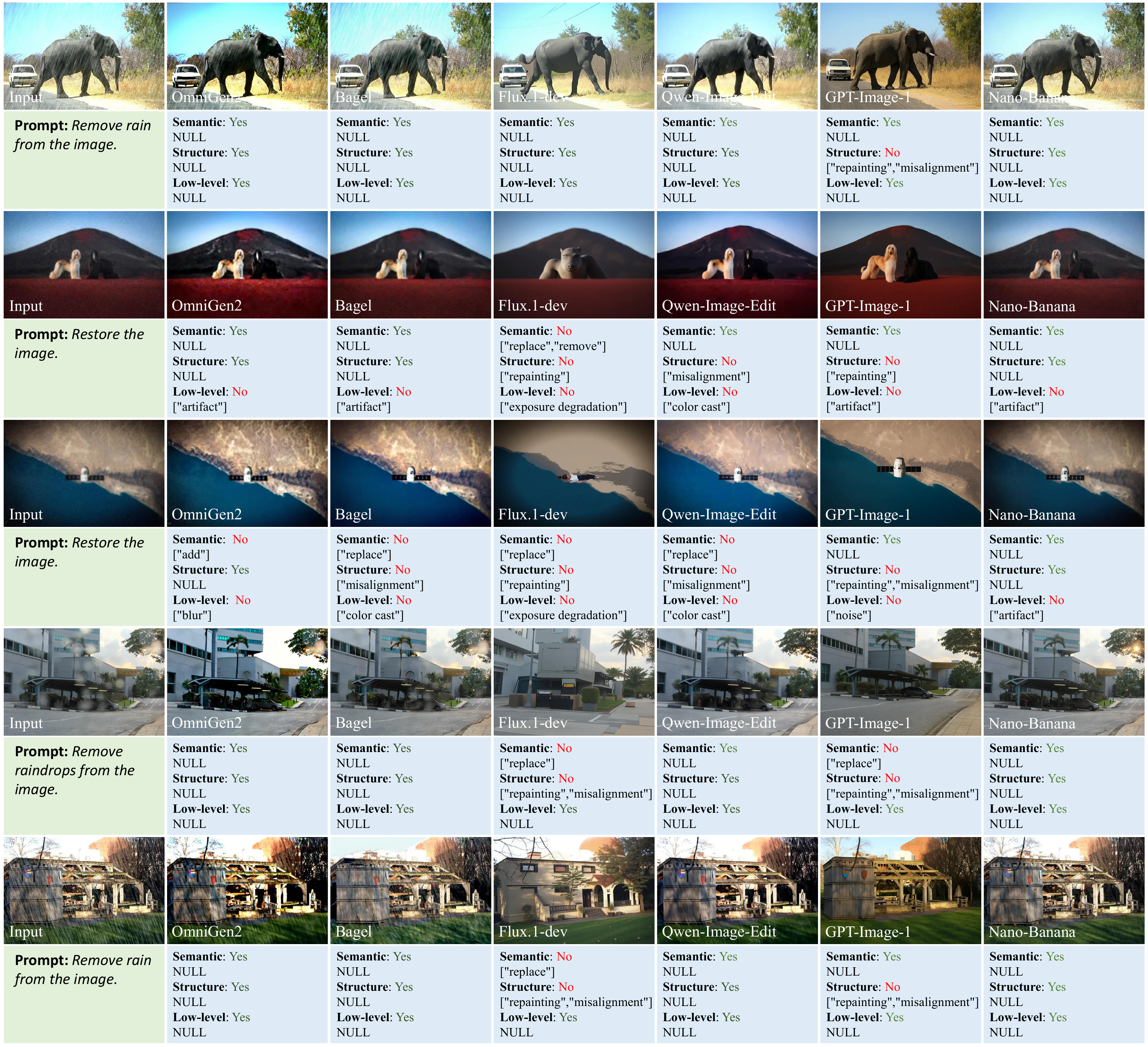} 
  \captionsetup{skip=1.5pt}
  \caption{Visualization of the evaluation results on Low-level Dataset using Format 1.}
  \vspace{-5pt}
  \label{fig:data_app_lowlevel}
\end{figure*}

To better illustrate the evaluation performance of our model, we present additional subjective result examples.
Specifically, the results on ImgEdit-Bench are shown in Fig.~\ref{fig:data_app_imgedit}, those on GEdit-Bench are shown in Fig.~\ref{fig:data_app_gedit}, and the results on the Low-level Dataset are shown in Fig.~\ref{fig:data_app_lowlevel}.

\begin{figure*}[h] 
\vspace{-5pt}
  \centering 
  \includegraphics[width=1\textwidth]{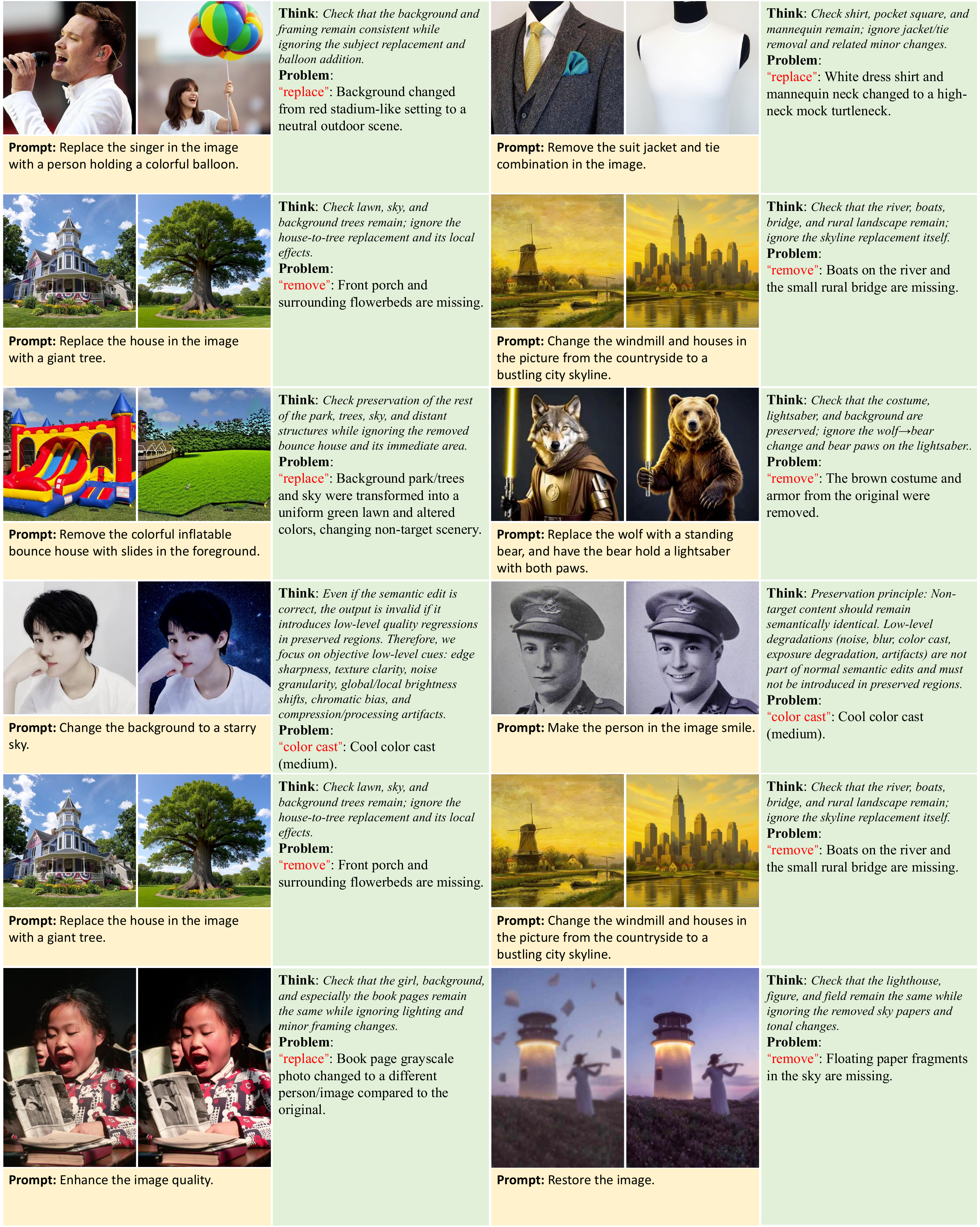} 
  \captionsetup{skip=1.5pt}
  \caption{Illustration of detailed results answered using Format 2.}
  \vspace{-5pt}
  \label{fig:data_app_des}
\end{figure*}

The results presented above correspond to the model’s short-form answers to the evaluation questions.
Fig.~\ref{fig:data_app_des} further illustrates detailed descriptions generated by the model for erroneous cases along the Semantic Level and Low-level Appearance dimensions.
In each example pair, the left image is the input image and the right image is the output image.
The model is able to accurately identify and describe both the error types and the corresponding target objects.

\subsection{Demonstration of Model Limitations}
\label{sec:B3}
\begin{figure*}[h] 
\vspace{-5pt}
  \centering 
  \includegraphics[width=1\textwidth]{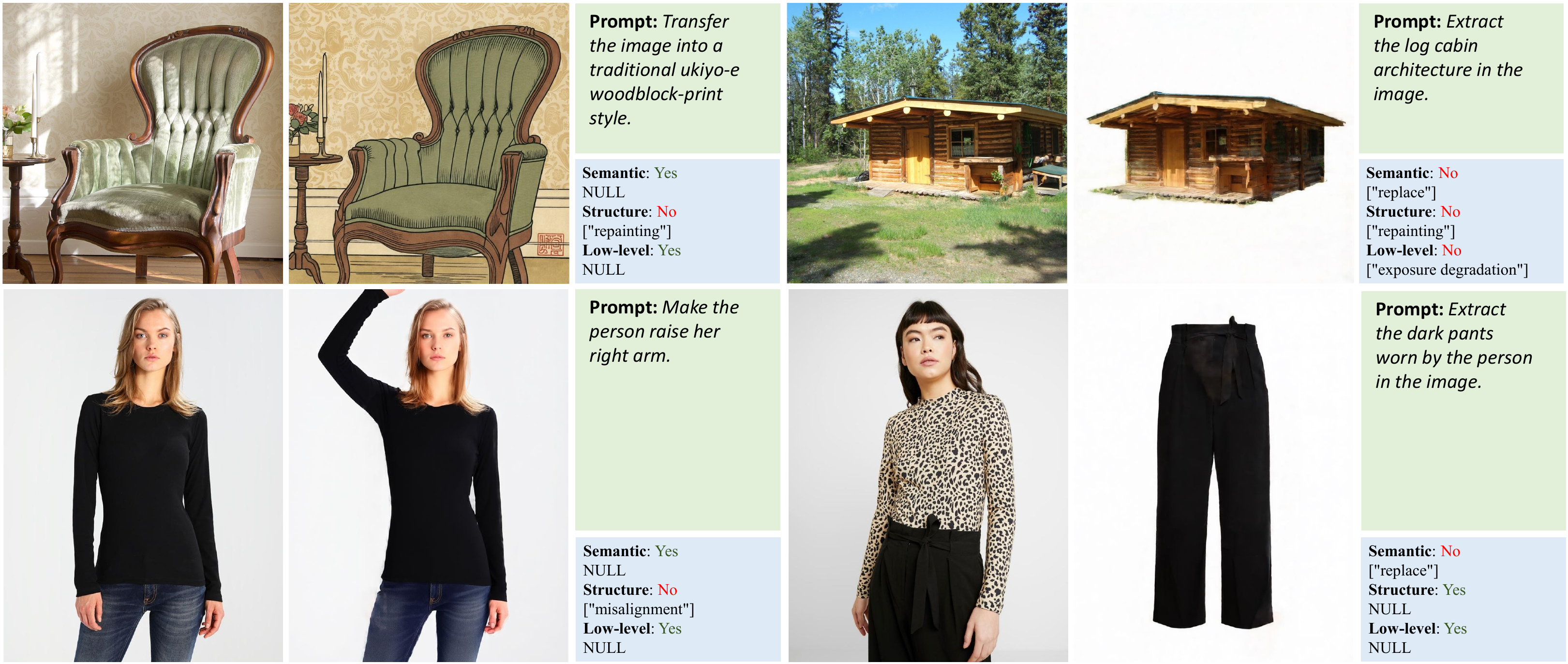} 
  \captionsetup{skip=1.5pt}
  \caption{The results shown above correspond to \textbf{failure cases} of StableI2I, reflecting its limitations.
Under human judgment, all of these tasks should be considered correct; however, the model fails to complete them successfully.
The specific editing types, from the top left to the bottom right, are: style transfer, object extraction, human motion, and object extraction.}
  \vspace{-5pt}
  \label{fig:app_limitation}
\end{figure*}

In our experiments, we also observed several failure cases, as shown in Fig.~\ref{fig:app_limitation}.
For example, in the case of style transfer, the style change itself constitutes a form of content repainting. From a human perspective, such an edit should be considered valid and correct; however, the model still classifies it as problematic repainting.
For object extraction tasks, the model may misjudge the result as losing other content, which can be regarded as a failure case that lies outside the distribution of the training data.
In addition, for human pose change scenarios, such edits inevitably cause pixel-level misalignment in other parts of the human body. Nevertheless, from a semantic and task-oriented perspective, these editing results are correct.

In summary, the errors made by the current model mainly stem from cases where the actual I2I processing direction overlaps or conflicts with the model’s judgment direction.
In the future, we plan to expand the scope of I2I tasks to better cover such cases, thereby enhancing the model’s perceptual capability for complex editing semantics.
Moreover, in such ambiguous situations, allowing the model to selectively abstain from answering can also be a reasonable solution.
We also believe that these issues are, to some extent, attributable to limitations in model capacity and parameter scale. In contrast, closed-source large models usually exhibit more flexible adaptation capabilities and tend to perform better in these scenarios.
If closed-source multimodal large models could achieve pixel-level perceptual sensitivity comparable to that of StableI2I, we believe that multimodal large models as a whole would make further progress.

\end{document}